\documentclass{article}

 \usepackage[preprint]{neurips_2022}

\usepackage{algorithm}
\usepackage{algpseudocode}
\usepackage[utf8]{inputenc} 
\usepackage[T1]{fontenc}    
\usepackage{hyperref}       
\usepackage{url}            
\usepackage{booktabs}       
\usepackage{amsfonts}       
\usepackage{nicefrac}       
\usepackage{microtype}      
\usepackage{xcolor}         
\usepackage{amssymb,amsmath,amscd,amsfonts,amsthm,bbm,mathrsfs,yhmath}

\usepackage{hyperref}
\usepackage{caption}

\usepackage{graphics}
\usepackage{mathtools}
\usepackage{cleveref}
\usepackage{comment}

\usepackage{paralist,enumitem}                                       
\usepackage{pdfpages}

\newtheorem{theorem}{Theorem}
\newtheorem{lemma}{Lemma}
\newtheorem{corollary}{Corollary}
\newtheorem{proposition}{Proposition}
\newtheorem{definition}{Definition}
\newtheorem{remark}{Remark}

\newtheorem{assumption}{Assumption}

\usepackage[textwidth=2cm, textsize=footnotesize]{todonotes}

\newcommand{\cO}{{\mathcal O}}

\newcommand{\cP}{{\mathcal P}}

\newcommand{\G}{{\textbf G}}

\newcommand{\E}{{{\mathbb E}}}
 
\newcommand{\R}{{\mathbb R}}

\newcommand{\cF}{\mathcal{F}}
\newcommand{\KL}[1]{H_{\pi}\left( #1\right)}
\newcommand{\FS}[1]{J_{\pi}\left( #1\right)}
\newcommand{\Exp}[1]{\mathbb{E}\left[ #1 \right]}
\newcommand{\norm}[1]{\left\| #1 \right\|}
\newcommand{\normsq}[1]{\left\| #1 \right\|^2}
\newcommand{\inner}[2]{\left< #1 , #2 \right>}
\newcommand{\dif}[1]{\frac{d#1}{dt}}
\newcommand{\algname}[1]{{\sf #1}}

\title{Federated Learning with a Sampling Algorithm\\ under Isoperimetry}

\author{
  Lukang Sun\\
  KAUST\\
  \\
  \texttt{lukang.sun@kaust.edu.sa} \\
  \And
  Adil Salim\\
  Microsoft Research\\
  \\
  \texttt{adilsalim@microsoft.com}
  \And
  Peter Richt\'{a}rik\\
  KAUST\\
  \\
\texttt{peter.richtarik@kaust.edu.sa}\\
}

\begin{document}

\maketitle

\begin{abstract}
	Federated learning uses a set of techniques to efficiently distribute the training of a machine learning algorithm across several devices, who own the training data. These techniques critically rely on reducing the communication cost---the main bottleneck---between the devices and a central server. Federated learning algorithms usually take an optimization approach: they are algorithms for minimizing the training loss subject to communication (and other) constraints. In this work, we instead take a Bayesian approach for the training task, and propose a communication-efficient variant of the Langevin algorithm to sample \textit{a posteriori}. The latter approach is more robust and provides more knowledge of the \textit{a posteriori} distribution than its optimization counterpart. We analyze our algorithm without assuming that the target distribution is strongly log-concave. Instead, we assume the weaker log Sobolev inequality, which allows for nonconvexity. 
\end{abstract}

\section{Introduction}

Federated learning uses a set of techniques to efficiently distribute the training of a machine learning algorithm across several devices, who own the training data. These techniques critically rely on reducing the communication cost---the main bottleneck---between the devices and a central server~\cite{konevcny2016federated,mcmahan2017communication}. Indeed, classical training algorithms such as the Stochastic Gradient Descent (SGD) can formally be distributed between the server and the devices. But this requires communications between the devices and the server, raising privacy and communication concerns. Therefore, most federated learning algorithms rely on compressing the messages that are exchanged between the server and the devices. For instance, variants SGD involving compression~\cite{karimireddy2020scaffold,horvath2019stochastic,haddadpour2021federated,das2020faster,Gorbunov2021} have successfully been applied in several engineering setups~\cite{kairouz2021advances,dimitriadis2022flute}.

More generally, most training algorithms in federated learning can be seen as communication efficient optimization algorithms minimizing the training loss $F$. This approach can provide an approximate minimizer, or a maximum \textit{a posteriori} estimator if we use the Bayesian language. However, this approach fails to provide sufficient knowledge of the \textit{a posteriori} distribution $\exp(-F)$ if we want to perform some Bayesian computation, such as computing confidence intervals. For Bayesian computation, samples from the \textit{a posteriori} distribution $\exp(-F)$ are preferred. In this case, the training task can be formalized as sampling from $\exp(-F)$, rather than minimizing $F$.

Exploiting the connections between optimization and sampling~\cite{wibisono2018sampling,durmus2019analysis}, we take a Bayesian approach to federated learning, similarly to~\cite{Vono2021,deng2021convergence,el2021federated}. We introduce a communication efficient variant of the Langevin algorithm, a widely used algorithm to sample from a target distribution whose density is proportional to $\exp(-F)$, where $F$ is a smooth nonconvex function. We study the complexity of our communication efficient algorithm to sample $\exp(-F)$. 

\subsection{Related Work}
Most approaches to train federated learning algorithms rely on minimizing the training loss with a communication efficient variant of SGD. Several training algorithms have been proposed, but we mention~\cite{karimireddy2020scaffold,horvath2019stochastic,haddadpour2021federated,das2020faster,Gorbunov2021} because these papers contain the state of the art results in terms of minimization of the training loss. In this work we shall specifically use the MARINA gradient estimator introduced in~\cite{Gorbunov2021} and inspired from~\cite{Nguyen2017}. 

The literature on sampling with Langevin algorithm is also large. In recent years, the machine learning research community has specifically been interested in the complexity of Langevin algorithm~\cite{dalalyan2017theoretical,durmus2017non}. In the case where $F$ is not convex, recent works include~\cite{vempala2019rapid,Wibisono2019,ma2019there,chewi2021} which use an isoperimetric-type inequality (\cite[Chapter 21]{villani2008}) such as the log Sobolev inequality to prove convergence in Kullback-Leibler divergence. Other works on Langevin in the nonconvex case include \cite{balasubramanian2022towards,cheng2018sharp,majka2020nonasymptotic,mattingly2002ergodicity}.

The closest papers to our work are~\cite{Vono2021,deng2021convergence}. Like our work, these papers are theoretical and study the convergence of an efficient variant of Langevin algorithm for federated learning. The key difference between these papers and our work is that they assume $F$ strongly convex, whereas we only assume that the target distribution $\pi \propto \exp(-F)$ satisfies the log Sobolev inequality (LSI). The strong convexity of $F$ implies LSI~\cite[Chapter 21]{villani2008}, but LSI allows $F$ to be nonconvex. We compare our complexity results to their in Table~\ref{tab:complexity}, Our contributions are summarized below.

\begin{table}[ht]
	\centering
\begin{tabular}{ |c|c|c|c| }
\hline
& & & \\
 Paper & Assumption & Criterion & Complexity \\ 
 & & & \\
 \hline
 & & & \\
\cite{Vono2021} & Strong convexity & $W_2(\rho_K,\pi)< \varepsilon$ & $K = \tilde{\mathcal O} \left(\frac{d}{\varepsilon^2}\right)$ \\
& & & \\
\hline
& & & \\
\cite{deng2021convergence} & Strong convexity & $W_2(\rho_K,\pi)< \varepsilon$ & $K = \tilde{\mathcal O} \left(\frac{d}{\varepsilon^2}\right)$ \\
& & & \\
\hline
& & & \\
\textbf{This paper} & \textbf{log Sobolev inequality} & $W_2(\rho_K,\pi)< \varepsilon$ & $K = \tilde{\mathcal O} \left(\frac{d}{\varepsilon^2}\right)$ \\
& & &  \\
\hline
\end{tabular}
\\
\caption{Sufficient number of iterations of our algorithm and concurrent algorithms to achieve $\varepsilon$ accuracy in 2-Wasserstein distance in dimension $d$. Strong convexity implies log Sobolev inequality.}
\label{tab:complexity}
\end{table}%

\subsection{Contributions}

We consider the problem of sampling from $\pi \propto \exp(-F)$, where $\pi$ satisfies LSI, in a federated learning setup. We make the following contributions:
\begin{itemize}
\item We propose a communication efficient variant of Langevin algorithm, called Langevin-MARINA, that can be distributively implemented between the central server and the devices. Langevin-MARINA relies on compressed communications.
\item We analyze the complexity of our sampling algorithm in terms of the Kullback Leibler divergence, the Total Variation distance and the 2-Wasserstein distance. Our approach relies on viewing our sampling problem as an optimization problem over a space of probability measures, and allows $F$ to be nonconvex.
\item Our sampling algorithm, Langevin-MARINA, is inspired from an optimization algorithm called MARINA~\cite{Gorbunov2021} for which we give a new convergence proof in the Appendix. This new proof draws connections between optimization (MARINA) and sampling (Langevin-MARINA).
\end{itemize}

\subsection{Paper organization}
In Section~\ref{sec:background}, we review some background material on sampling and optimization. Next, we introduce our federated learning setup in Section~\ref{sec:fl}. In Section~\ref{sec:main} we give our main algorithm, Langevin-MARINA, and our main complexity results. We conclude in Section~\ref{sec:ccl}. The proofs, including the new proofs of existing results for MARINA, are deferred to the Appendix. 
%

	\section{Preliminaries}\label{sec:background}
	\subsection{Mathematical problem}
	Throughout this paper, we consider a nonconvex function $F: \R^d \to \R$ which is assumed to be $L$-smooth, i.e., differentiable with an $L$-Lipschitz continuous gradient: $\norm{\nabla F(x)-\nabla F(y)}\leq L\norm{x-y}$. As often in machine learning, $F$ can be seen as a training loss and takes the form of a finite sum 
	\begin{equation}
	    \label{eq:finitesum}
	    F = \sum_{i = 1}^n F_i,
	\end{equation}
	where each $F_i$ can be seen as the loss associated to the dataset stored in the device $i$. Assuming that $\int \exp(-F(x))dx \in (0,\infty)$, we denote by 
	\begin{equation}
	\label{eq:sampling}
	    \pi \propto \exp(-F),
	\end{equation} the probability distribution whose density is proportional to $\exp(-F)$. We take a Bayesian approach: instead of minimizing $F$, our goal is to generate random samples from $\pi$, which can be seen as the posterior distribution of some Bayesian model.
	

\subsection{Sampling and Optimal Transport}
We denote by $\cP_2(\R^d)$ the set of Borel measures $\sigma$ on $\R^d$ with finite second moment, that is  $\int\normsq{x}d\sigma(x)<+\infty$. For every $\sigma, \nu\in\cP_2(\R^d)$, $\Gamma(\sigma,\nu)$ is the set of all the coupling measures between $\sigma$ and $\nu$ on $\R^{d}\times\R^d$, that is $\gamma\in\Gamma(\sigma,\nu)$ if and only if $\sigma(dx)= \gamma(dx,\R^d)$ and $\nu(dy)= \gamma(\R^d,dy)$. The Wasserstein distance between $\sigma$ and $\nu$ is defined by
\begin{equation}
	\label{eq:wstdis}
	W_2(\sigma,\nu)=\sqrt{\inf_{\gamma\in\Gamma(\sigma,\nu)}\int \normsq{x-y}\gamma(dx,dy)}.
\end{equation}
The Wasserstein distance $W_2(\cdot,\cdot)$ is a metric on $\cP_2(\R^d)$ and the metric space $\left(\cP_2(\R^d),W_2\right)$ is called the Wasserstein space, see \cite{ambrosio2008gradient}. 



The Kullback-Leibler~(KL) divergence w.r.t. $\pi$ can be seen as the map from the Wasserstein space to $(0,\infty]$ defined for every $\sigma \in \cP_2(\R^d)$ by
\begin{equation}
	\label{eq:KLdiv}
	\KL{\sigma}:=\begin{cases}
		\int\log(\frac{\sigma}{\pi})(x)d\sigma(x) & \text{if $\sigma \ll \pi$} \\
		\infty                              & \text{else},
	\end{cases}
\end{equation}
where $\sigma \ll \pi$ means that $\sigma$ is absolutely continuous w.r.t. $\pi$ and $\frac{\sigma}{\pi}$ denotes the density of $\sigma$ w.r.t. $\pi$.
The KL divergence is always nonnegative and it is equal to zero if and only if $\sigma = \pi$. Therefore, assuming that $\pi \in \cP_2(\R^d)$, $\pi$ can be seen as the solution to the optimization problem
\begin{equation}
\label{eq:optim-sampling}
    \min_{\sigma \in \cP_2(\R^d)} \KL{\sigma}.
\end{equation}
Viewing $\pi$ as a minimizer of the KL divergence is the cornerstone of our approach. Indeed, we shall view the proposed algorithm as an optimization algorithm to solve Problem~\eqref{eq:optim-sampling}. In particular, following~\cite{wibisono2018sampling,durmus2019analysis}, we shall view the Langevin algorithm as a first order algorithm over the Wasserstein space. In particular, the relative Fisher information, defined as
\begin{equation}
	\label{eq:fishinfor}
	\FS{\sigma}:=\begin{cases}
		\int \normsq{\nabla \log(\frac{\sigma}{\pi})(x)}d\sigma(x)&\text{if $\sigma  \ll \pi$} \\
		\infty,                             & \text{else},
	\end{cases}
\end{equation}
will play the role of the squared norm of the gradient of the objective function. Indeed, by defining a differential structure over the Wasserstein space~\cite{ambrosio2008gradient}, the relative Fisher information is the squared norm of the gradient of the KL divergence.

The analysis of nonconvex optimization algorithm minimizing $F$ often relies on a gradient domination condition relating the squared norm of the gradient to the function values, such as the following Lojasiewicz condition~\cite{blanchet2018family,karimi2016linear}
\begin{equation}
		\label{eq:PL}
		F(x)- \min F \leq \frac{1}{2\mu}\|\nabla F(x)\|^{2}.
	\end{equation}
The latter condition has a well-known analogue over the Wasserstein space for the KL divergence, called logarithmic Sobolev inequality.
\begin{assumption}[Logarithmic Sobolev Inequality~(LSI)]
	\label{def:LSI}
	The distribution $\pi$ satisfies the logarithmic Sobolev inequality with constant $\mu$: for all $\sigma\in\cP_2(\R^d)$, 
		\begin{equation}
		\label{eq:LSIinequ}
		\KL{\sigma}\leq\frac{1}{2\mu}\FS{\sigma}.
	\end{equation}
\end{assumption}
The log Sobolev inequality has been studied in the optimal transport community~\cite[Chapter 21]{villani2008} and holds for several target distributions $\pi$. First, if $F$ is $\mu$-strongly convex (we say that $\pi$ is strongly log-concave), then LSI holds with constant $\mu$. Besides, LSI is preserved under bounded perturbations and Lipschitz mappings~\cite[Lemma 16, 19]{vempala2019rapid}. Therefore, small perturbations of strongly-log concave distributions satisfy LSI. For instance, if we add to a Gaussian distribution another Gaussian distribution with a small weight, then the resulting mixture of Gaussian distributions is not log-concave but it satisfies LSI. One can also find compactly supported examples, see~\cite[Introduction]{Wibisono2019}.

LSI is a condition that has been used in the analysis of the Langevin algorithm
\begin{equation}
\label{eq:langevin}
	x_{k+1}=x_k-h\nabla F(x_k)+\sqrt{2h}Z_{k+1},
\end{equation}
where $h>0$ is a step size and $(Z_k)$ a sequence of i.i.d standard Gaussian vectors over $\R^d$. Langevin algorithm can be seen as a gradient descent algorithm for $F$ to which a Gaussian noise is added at each step. For instance, \citep{Wibisono2019} showed that the distributions $\rho_k$ of $x_k$ converges rapidly towards the target distribution $\pi$ in terms of the KL divergence under LSI. Other metrics have also been considered, such a the Total Variation distance 
\begin{equation}
    \label{eq:TV}
    \|\sigma-\pi\|_{TV} = \sup_{A \in B(\R^d)}|\sigma(A) - \pi(A)|,
\end{equation}
where $B(\R^d)$ denotes the Borel sigma field of $\R^d$.

\section{Federated learning}
\label{sec:fl}
\subsection{Example of a federated learning algorithm: MARINA}

We now describe our federated learning setup. A central server and a number of devices are required to run a training algorithm in a distributed manner. Each device $i \in \{1,\ldots,n\}$ owns a dataset. A meta federated learning algorithm is as follows: at each iteration, each device performs some local computation (for instance, a gradient computation) using its own dataset, then each device sends the result of that computation to the central server. The central server aggregates the results and sends that aggregation back to the devices. 
Algorithm~\ref{alg:capmarina} provides an example of a federated learning algorithm.
\begin{algorithm}[t!]
	\caption{\algname{MARINA}~\cite{Gorbunov2021}}\label{alg:capmarina}
	\begin{algorithmic}[1]
		\State {\bfseries Input:} Starting point $x_0$, step-size $h$, number of iterations $K$
		\For{$i=1,2,\cdots,n$ in parallel}
		\State Device $i$ computes MARINA estimator $g_{0}^i$ 
		\State Device $i$ uploads $g_0^i$ to the central server 
		\EndFor
		\State Server aggregates $g_0=\frac{1}{n}\sum_{i=1}^ng_0^i$
		\For {$k=0,1,2,\cdots,K-1$}
		\State Server broadcasts $g_k$ to all devices $i$
		\For{$i=1,2,\cdots,n$ in parallel}
		\State Device $i$ performs $x_{k+1}=x_k-h g_k$
		\State Device $i$ computes MARINA estiamtor $g_{k+1}^i$
		\State Device $i$ uploads $g_{k+1}^i$ to the central server
		\EndFor
		\State Server aggregates $g_{k+1}=\frac{1}{n}\sum_{i=1}^ng_{k+1}^i$
		\EndFor
		\State {\bfseries Return:} $\hat{x}^K$ chosen uniformly at random from $\left(x_k\right)_{k=0}^K$ or last point $x_K$
	\end{algorithmic}
\end{algorithm}
This algorithm, called MARINA, was introduced and analyzed in~\cite{Gorbunov2021} under the assumption that the sequence $(g_k)$ is a MARINA estimator of the gradient as defined below.
\begin{definition}
	\label{def:marina}
	Given a sequence produced by an algorithm $(x_k)_k$, a MARINA estimator of the gradient is a random sequence $(g_k)_k$ satisfying $\Exp{g_k} = \Exp{\nabla F(x_k)}$ and
	\begin{equation}
		\label{eq:LAboundg111}
		\G_{k+1}\leq (1-p)\G_k+(1-p)L^2\alpha\Exp{\normsq{x_{k+1}-x_k}}+\theta,
	\end{equation}
	where $\G_k = \Exp{\normsq{g_k-\nabla F(x_k)}}$ and $0 < p \leq 1$, $\alpha,\theta \geq 0$. 
\end{definition}
MARINA iterations can be rewritten as
\begin{equation}
\label{eq:MARINAit}
    x_{k+1} = x_k - h g_{k}.
\end{equation}
Since $\Exp{g_k} = \Exp{\nabla F(x_k)}$, MARINA can be seen as a SGD distributed between the central server and the devices, to minimize the nonconvex function $F$.

Recall that in federated learning, the main bottleneck is the communication cost between server and devices. The key aspect of MARINA is that it achieves the state of the art in terms of communication complexity among nonconvex optimization algorithms. The reason is the following: there are practical examples of MARINA estimators $(g_k)$ that rely on a compression step before the communication and are therefore cheap to communicate\footnote{Note that we did not specify how $g_{k}^i$ is computed in Algorithm~\ref{alg:capmarina}. For the moment we only require $g_k$ to be a MARINA estimator}.

\subsection{Compression operator}
MARINA estimators are typically cheaper to communicate than full gradient because they involve a compression step, making the communication light. A compression operator is a random map from $\R^d$ to $\R^d$ with the following properties.
\begin{definition}[Compression]
	\label{def:compression}
	A stochastic mapping $\mathcal{Q}: \mathbb{R}^{d} \rightarrow \mathbb{R}^{d}$ is a compression operator if there exists $\omega>0$ such that for any $x \in \mathbb{R}^{d}$,
	\begin{equation}
		\label{eq:sgskjf}
		\Exp{\mathcal{Q}(x)}=x, \quad \Exp{\|\mathcal{Q}(x)-x\|^{2}} \leq \omega\|x\|^{2}.
	\end{equation}
\end{definition}
The first equation states that $\mathcal{Q}$ is unbiased and the second equation states that the variance of the compression operator has a quadratic growth. There are lots of compression operators satisfying \eqref{eq:sgskjf}, explicit examples based on quantization and/or sparsification are given in~\cite{alistarh2017qsgd,horvath2019natural}. In general, compressed quantities are cheap to communicate.

We now fix a compression operator and give examples of gradient estimators involving compression, which are provably MARINA estimators.




\subsection{\algname{MARINA} gradient estimators}\label{subsec:11}

In every examples presented below, $g_k:=\frac{1}{n}\sum_{i=1}^n g_k^i$, where $g_k^i$ is meant to approximate $\nabla F_i(x_k)$, i.e., the gradient at the device $i$, at step $k$. We consider any random sequence $(x_k)$ generated by a training algorithm.



\subsubsection{Vanilla MARINA gradient estimator}
The vanilla MARINA gradient estimator is defined as the average $g_k:=\frac{1}{n}\sum_{i=1}^n g_k^i$, where for every $i \in \{1,\dots,n\}$, $g_0^i=\nabla F_i(x_0)$ and 
\begin{equation}
	\label{eq:gradientestimator}
	g^i_{k+1}:=\begin{cases}
		\nabla F_i(x_{k+1})& \text{with probability }~ p>0~\\
		g_k+\mathcal{Q}\left(\nabla F_i(x_{k+1})-\nabla F_i(x_k)\right)& \text{with probability }~1-p
	\end{cases}.
\end{equation}

Node $i$ randomly (and independently of $\mathcal{Q}, g_k, x_{k+1}$) computes and uploads to the server the gradient of $F_i$ at point $x_{k+1}$ with probability $p$. Else, node $i$ computes and uploads the compressed difference of the gradients of $F_i$ between points $x_{k+1}$ and $x_{k}$: $\mathcal{Q}\left(\nabla F_i(x_{k+1})-\nabla F_i(x_k)\right)$. Note that the server already knows $g_k$ from the previous iteration, so no need to send $g_k$ to the server. In average, device $i$ computes a full gradient every $1/p$ steps ($p$ is very small, see Remark~\ref{rk} below).

\begin{proposition}
	\label{prop:vanilla}
	Assume that $F_i$ is $L_i$-smooth for all $i\in \{1,\ldots,n\}$, i.e.,
	\begin{equation}\label{eq:Lismooth}
		\left\|\nabla F_{i}(x)-\nabla F_{i}(y)\right\| \leq L_{i}\|x-y\|,\quad \forall x, y \in \mathbb{R}^{d}, \forall i\in \{1,\ldots,n\}.
	\end{equation}
	Then, $g_k = \frac{1}{n}\sum_{i=1}^n g_k^i$, where $g_k^i$ is defined by~\eqref{eq:gradientestimator} is a MARINA estimator in the sense of Definition~\ref{def:marina}, with $\alpha=\frac{\omega\sum_{i=1}^n L_i^2}{n^2L^2},\theta=0$.
\end{proposition}
This proposition means that the MARINA estimator reduces the variance induced by the compression operator. Note that, if $F_i$ is $L_i$-smooth, then $L \leq \frac{1}{n} \sum_{i=1}^n L_i$.





\subsubsection{Finite sum case}
Consider the finite sum case where each $F_i$ is a sum over the data points stored in the device $i$: $F_i = \sum_{j = 1}^N F_{ij}$, where $F_{ij}$ is the loss associated to the data point $j$ in the device $i$.
The finite sum MARINA gradient estimator is analogue to the vanilla estimator, but with subsampling. It is defined as the average $g_k:=\frac{1}{n}\sum_{i=1}^n g_k^i$, where for every $i \in \{1,\dots,n\}$, $g_0^i=\nabla F_i(x_0)$ and 
\begin{equation}
	\label{eq:gradientestimator222}
	g^i_{k+1}:=\begin{cases}
		\nabla F_i(x_{k+1})& \text{with probability }~ p>0~\\
		g_k+\mathcal{Q}\left(\frac{1}{b'}\sum_{j\in I_{i,k}^{\prime}}\left(\nabla F_{ij}(x_{k+1})-\nabla F_{ij}(x_k)\right)\right)& \text{with probability }~1-p
	\end{cases},
\end{equation}
where $b'$ is the minibatch size and $I_{i, k}^{\prime}$ is the set of the indices in the minibatch, $\left|I_{i, k}^{\prime}\right|=b^{\prime}$. Here, $(I_{i, k}^{\prime})_{i,k}$ are i.i.d random sets consisting in $b'$ i.i.d samples from the uniform distribution over $\{1,\ldots,N\}$.

\begin{proposition}
	\label{prop:finitesum}
	Assume that $F_{ij}$ is $L_{ij}$-smooth for all $i \in \{1,\ldots,n\}$ and all $j \in \{1,\ldots,N\}$. 
	Then, $g_k = \frac{1}{n}\sum_{i=1}^n g_k^i$, where $g_k^i$ is defined by~\eqref{eq:gradientestimator222} is a MARINA estimator in the sense of Definition~\ref{def:marina}, with \[\alpha=\frac{\omega\sum_{i=1}^nL_i^2+(1+\omega)\frac{\sum_{i=1}^n\mathcal{L}^2_i}{b'}}{n^2L^2},\]where $\mathcal{L}_{i} = \max _{j \in[N]} L_{i j}$ and $\theta=0$.
\end{proposition}
This proposition means that the MARINA estimator reduces the variance induced by the compression operator and the variance induced by the subsampling.

\subsubsection{Online case}
\label{sec:online}
Consider the online case where each $F_i$ is an expectation over the randomness of a stream of data arriving online: $F_{i}(x)=\mathbb{E}_{\xi_{i} \sim \mathcal{D}_{i}}\left[F_{\xi_{i}}(x)\right]$, where $F_{\xi_{i}}$ is the loss associated to a data point $\xi_i$ in device $i$. The online MARINA gradient estimator is analogue to the finite sum estimator, but with subsampling from $\mathcal{D}_{i}$ instead of the uniform over $\{1,\ldots,N\}$. Moreover, the full gradient $\nabla F_i$ is never computed (because it is intractable in online learning). The online MARINA gradient estimator is defined as the average $g_k:=\frac{1}{n}\sum_{i=1}^n g_k^i$, where for every $i \in \{1,\dots,n\}$, $g_0^i=\frac{1}{b} \sum_{\xi \in I_{i, 0}} \nabla F_{\xi}\left(x_{0}\right)$ and
\begin{equation}
	\label{eq:gradientestimator333}
	g^i_{k+1}:=\begin{cases}
		\frac{1}{b} \sum_{\xi \in I_{i, k}} \nabla F_{\xi}\left(x_{k+1}\right)& \text{with prob.}~ p>0~\\
		g_k+\mathcal{Q}\left(\frac{1}{b^{\prime}} \sum_{\xi \in I_{i, k}^{\prime}}\left(\nabla F_{\xi}\left(x_{k+1}\right)-\nabla F_{\xi}\left(x_{k}\right)\right)\right)& \text{with prob.} ~1-p
	\end{cases},
\end{equation}
where $I_{i, k}^{\prime}$ and $I_{i, k}$ are the set of the indices in a minibatch, $\left|I_{i, k}^{\prime}\right|=b^{\prime}$ and $\left|I_{i, k}\right|=b$. Here, $(I_{i, k})_{i,k}$ (resp. $(I_{i, k}^{\prime})_{i,k}$) is an i.i.d random set consisting in $b$ (resp. $b'$) i.i.d samples from $\mathcal{D}_{i}$. Besides, in the online case only, we make a bounded gradient assumption.
	\begin{proposition}
		\label{prop:online}
		Assume that for all $i \in\{1,\ldots,n\}$ there exists $\sigma_{i} \geq 0$ such that for all $x \in \mathbb{R}^{d}$,
		\begin{align*}
		&\mathbb{E}_{\xi_{i} \sim \mathcal{D}_{i}}\left[\nabla F_{\xi_{i}}(x)\right] =\nabla F_{i}(x), \\
		&\mathbb{E}_{\xi_{i} \sim \mathcal{D}_{i}}\left[\left\|\nabla F_{\xi_{i}}(x)-\nabla F_{i}(x)\right\|^{2}\right] \leq \sigma_{i}^{2}.
	\end{align*}
		Moreover, assume that $F_{\xi_i}$ is $L_{\xi_i}$-smooth $\mathcal D_i$ almost surely. 
		Then, $g_k = \frac{1}{n}\sum_{i=1}^n g_k^i$, where $g_k^i$ is defined by~\eqref{eq:gradientestimator333} is a MARINA estimator in the sense of Definition~\ref{def:marina}, with \[\alpha=\frac{\omega\sum_{i=1}^nL_i^2+(1+\omega)\frac{\sum_{i=1}^n\mathcal{L}^2_i}{b'}}{n^2L^2},\] where $\mathcal{L}_{i}$ is the $L^\infty$ norm of $L_{\xi_i}$ (where $\xi_i \sim \mathcal D_i$) and $\theta=\frac{p\sum_{i=1}^n\sigma_i^2}{n^2b}$.
	\end{proposition}
	Note that in the online case $\theta \neq 0$ in general, because the online MARINA estimator does not reduce to zero the variance induced by the subsampling. That is because the loss $F_i$ is an expectation (and not a finite sum \textit{a priori}), for which variance reduction to zero is impossible in general. However, $\theta$ can be made small by taking a large minibatch size $b$ or a small probability $p$. Moreover, in the online case, $\G_0 \neq 0$ unlike in the two other cases.
	
	\begin{remark}
		\label{rk}
		Typically, $p$ is very small, for instance, \cite{Gorbunov2021} chooses $p=\zeta_{\mathcal{Q}}/d$ in \eqref{eq:gradientestimator}, $p=\min\left\{\zeta_{\mathcal{Q}}/d,b'/(N+b')\right\}$ in \eqref{eq:gradientestimator222} and $p=\min\left\{\zeta_{\mathcal{Q}}/d,b'/(b+b')\right\}$ in \eqref{eq:gradientestimator333}, where  $\zeta_{\mathcal{Q}}=\sup _{x \in \mathbb{R}^{d}} \mathbb{E}\left[\|\mathcal{Q}(x)\|_{0}\right]$ and $\|y\|_{0}$ is the number of non-zero components of $y \in \mathbb{R}^{d}$. Therefore, with high probability $1-p$, the compressed difference $\mathcal{Q}\left(\nabla F_i(x_{k+1})-\nabla F_i(x_k)\right)$ is sent to the server. Sending compressed quantities has a low communication cost~\cite{alistarh2017qsgd,horvath2019natural}. 
	\end{remark}
	
\begin{remark}
    Propositions~\ref{prop:vanilla}, \ref{prop:finitesum} and~\ref{prop:online} can be found in~\cite[Equation 21, 33, 46]{Gorbunov2021} in the case where $(x_k)$ is a stochastic gradient descent algorithm (i.e., $x_{k+1} = x_k - h g^k$). In the general case where $(x_k)$ is produced by any algorithm, we found out that the proofs of these Propositions are the same, but we reproduce these proofs in the Appendix for the sake of completeness.

\end{remark}

\section{Langevin-MARINA}\label{sec:main}
In this section, we give our main algorithm, Langevin-MARINA, and study its convergence for sampling from $\pi \propto \exp(-F)$. We prove convergence bounds in KL divergence, 2-Wasserstein distance and Total Variation distance under LSI. 




Our main algorithm, Langevin-MARINA can be seen as a Langevin variant of MARINA which adds a Gaussian noise at each step of MARINA (Algorithm~\ref{alg:capmarina}). Our motivation is to obtain a Langevin algorithm whose communication complexity is similar to that of MARINA. Alternatively, one can see Langevin-MARINA as a MARINA variant of Langevin algorithm which uses a MARINA estimator of the gradient $g_k=\frac{1}{n}\sum_{i=1}^n g_k^i$ instead of the full gradient $\nabla F(x_k)$ in the Langevin algorithm:
\begin{equation}
\label{eq:LMARINAit}
    x_{k+1} = x_k - h g_{k} + \sqrt{2h}Z_{k+1}.
\end{equation}

Langevin-MARINA is presented in Algorithm~\ref{alg:federatedsampling}.

\begin{algorithm}[h!]
	\caption{Langevin-MARINA (proposed algorithm)}\label{alg:federatedsampling}
	\begin{algorithmic}[1]
		\State {\bfseries Input:} Starting point $x_0\sim\rho_0$, step-size $h$, number of iterations $K$
		\For{$i=1,2,\cdots,n$ in parallel}
		\State Device $i$ computes MARINA estimator $g_{0}^i$ 
		\State Device $i$ uploads $g_0^i$ to the central server 
		\EndFor
		\State Server aggregates $g_0=\frac{1}{n}\sum_{i=1}^ng_0^i$
		\For {$k=0,1,2,\cdots,K-1$}
		\State Server broadcasts $g_k$ to all devices $i$
		\State Server draws a Gaussian vector $Z_{k+1}\sim\mathcal{N}(0,I_{d})$
		\State Server performs $x_{k+1}=x_k-hg_k+\sqrt{2h}Z_{k+1}$
		\State Server broadcasts $ x_{k+1}$ to all devices $i$
		\For{$i=1,2,\cdots,n$ in parallel}
		\State Device $i$ computes MARINA estimator $g_{k+1}^i$
		\State Device $i$ uploads $g^i_{k+1}$ to the central server
		\EndFor
		\State Server aggregates $g_{k+1}=\frac{1}{n}\sum_{i=1}^ng_{k+1}^i$ 
		\EndFor
		\State {\bfseries Return:} $x_K$
	\end{algorithmic}
\end{algorithm}

Compared to MARINA where $n$ equivalent stochastic gradient descent steps are performed by the devices, here the Langevin step is performed only once, by the server. This allows for computation savings ($n$ times less computations) at the cost of more communication: the iterates $x_k$ need to be broadcast by the server to the devices. Therefore, the communication complexity of the sampling algorithm Langevin-MARINA is higher than the communication complexity of the optimization algorithm MARINA. 

However, compared to the concurrent sampling algorithm of~\cite{Vono2021}, the communication complexity per iteration of Langevin-MARINA is equivalent. Comparing the communication complexity of Langevin-MARINA to that of FA-LD~\cite{deng2021convergence} is more difficult because FA-LD makes communication savings by performing communication rounds only after a number $T$ of local updates, instead of using a compression operator after each local update like Langevin-MARINA.


We first prove the convergence of Langevin-MARINA in KL-divergence. 

\begin{theorem}
	\label{thm:federatedsampling}
	Assume that LSI (Assumption~\ref{def:LSI}) holds and that $g_k = \frac{1}{n}\sum_{i=1}^n g_k^i$ is a MARINA estimator in the sense of Definition~\ref{def:marina}. If
		\begin{equation}\label{eq:sdksjfsk}
		0<h\leq \min \left\{\frac{1}{14L}\sqrt{\frac{p}{1+\alpha}},\frac{p}{6\mu} \right\},
	\end{equation}
	then
	\begin{equation}\label{eq:KLre}
		\KL{\rho_{K}}\leq e^{-\mu K h}\Psi_3+\frac{1-e^{-K\mu h}}{\mu}\tau,
	\end{equation}
	where  $\Psi_3=\KL{\rho_0}+\frac{1-e^{-\mu h}}{\mu}C\G_{0}, \tau=\left(2L^2+C(1-p)L^2\alpha\right)\left(8Lh^2d+4dh\right)+C\theta, C=\frac{8L^2h^2\beta+2\beta}{1-(1-p)\left(4L^2h^2\alpha+1\right)\beta}, \beta=e^{\mu h}$ and $\rho_k$ is the distribution of $x_k$ for every $k$.
	
	In particular, if $\theta=0$, set $h=\mathcal{O}\left(\frac{\mu p\varepsilon}{L^2(1+\alpha) d}\right)$,  $K=\Omega\left(\frac{L^2(1+\alpha) d}{\mu^2 p\varepsilon}\log \left(\frac{\Psi_3}{\varepsilon}\right)\right)$  then $\KL{\rho_{K}} \leq \varepsilon$.
	If $\theta\neq 0$, there will always be an extra residual term $\frac{1-e^{-K\mu h}}{\mu}C\theta$ in the right hand side of \eqref{eq:KLre} which cannot be diminished to 0 by setting $K$ and $h$. However in the online case (Section~\ref{sec:online}), $\theta=\frac{p\sum_{i=1}^n\sigma_i^2}{n^2b}$, we can set $b=\Omega\left(\frac{\sum_{i=1}^{n} \sigma_{i}^{2}}{\mu n^{2} \varepsilon}\right)$ to make the residual term small $\frac{1-e^{-K\mu h}}{\mu}C\theta=\cO(\varepsilon)$.

\end{theorem}
From this theorem, we can obtain complexity results in Total Variation distance and 2-Wasserstein distance. Indeed, Pinsker inequality states that 
\begin{equation}
	\|\sigma-\pi\|_{TV}\leq\sqrt{\frac{1}{2}H_{\pi}{(\sigma)}},\forall
	\sigma,\nu\in\mathcal{P}_2(\R^d),
\end{equation}
and LSI implies Talagrand's $T_2$ inequality
\begin{equation}
	W_2^2(\sigma,\pi)\leq  \frac{2}{\mu}\KL{\sigma},\forall\sigma\in\mathcal{P}_2(\R^d),
\end{equation}
 see \cite[Chapter 21]{villani2008}.
\begin{corollary}\label{cor:12}
Let assumptions and parameters be as in Theorem \ref{thm:federatedsampling}. Then, 
		\begin{equation}\label{eq:totrm}
		\|\rho_K-\pi\|^2_{TV}\leq  \frac{1}{2}\left(e^{-\mu K h}\Psi_3+\frac{1-e^{-K\mu h}}{\mu}\tau\right),
	\end{equation}
	and 
		\begin{equation}\label{eq:cor-W2}
		W^2_2(\rho_K,\pi)\leq \frac{2}{\mu}\left(e^{-\mu K h}\Psi_3+\frac{1-e^{-K\mu h}}{\mu}\tau\right).
	\end{equation}
\end{corollary}

In particular, if $\theta = 0$, set $h=\mathcal{O}\left(\frac{\mu
	p\varepsilon^2}{L^2(1+\alpha) d}\right)$,  $K=\Omega\left(\frac{L^2(1+\alpha) d}{\mu^2
	p\varepsilon^2}\log\left(\frac{\Psi_3}{\varepsilon^2}\right)\right)$ 
 then $\|\rho_K-\pi\|_{TV}\leq \varepsilon$. 
 	If $\theta\neq 0$, there will always be an extra residual term $\frac{1-e^{-K\mu h}}{\mu}C\theta$ in the right hand side of \eqref{eq:totrm} which cannot be diminished to 0 by setting $K$ and $h$. However in the online case (Section~\ref{sec:online}), $\theta=\frac{p\sum_{i=1}^n\sigma_i^2}{n^2b}$, we can set $b=\Omega\left(\frac{\sum_{i=1}^{n} \sigma_{i}^{2}}{\mu n^{2} \varepsilon^2}\right)$ to make the residual term small $\frac{1-e^{-K\mu h}}{\mu}C\theta=\cO(\varepsilon^2)$.

Finally, if $\theta=0$, set $h=\mathcal{O}\left(\frac{\mu^2 p\varepsilon^2}{L^2(1+\alpha) d}\right)$, $K=\Omega\left(\frac{L^2(1+\alpha) d}{\mu^3 p\varepsilon^2}\log\left(\frac{\Psi_3}{\mu\varepsilon^2}\right)\right)$ then $W_2(\rho_K,\pi)\leq\varepsilon$. 
	If $\theta\neq 0$, there will always be an extra residual term $\frac{1-e^{-K\mu h}}{\mu^2}C\theta$ in the right hand side of \eqref{eq:cor-W2} which cannot be diminished to 0 by setting $K$ and $h$. However in the online case, $\theta=\frac{p\sum_{i=1}^n\sigma_i^2}{n^2b}$, we can set $b=\Omega\left(\frac{\sum_{i=1}^{n} \sigma_{i}^{2}}{\mu^2 n^{2} \varepsilon^2}\right)$ to make the residual term small $\frac{1-e^{-K\mu h}}{\mu^2}C\theta=\cO(\varepsilon^2)$.

We can compare our results to~\cite{Vono2021,deng2021convergence} which also obtain bounds in 2-Wasserstein distance under the stronger assumption that $F$ is $\mu$-strongly convex, see Table~\ref{tab:complexity}.
In \cite{Vono2021}, they need $h=\cO\left(\frac{\varepsilon^2}{dl}\right)$ and $K=\Omega\left(\frac{dl}{\varepsilon^2}\log\left(\frac{W_2^2(\rho_0,\pi)}{\varepsilon^2}\right)\right)$
, where they compute the full gradient every $l$ step ~(we do not include the dependence in $\mu,L$ in their result for the sake of simplicity). One can think of $l$ as $1/p$.
In \cite{deng2021convergence}, they require $h=\cO\left(\frac{\mu^2\varepsilon^2}{dL^2\left(T^2\mu+L\right)}\right)$ and $K=\Omega\left(\frac{dL^2\left(T^2\mu+L\right)}{\mu^3\varepsilon^2}\log\left(\frac{d}{\varepsilon^2}\right)\right)$ to get $W_2(\rho_K,\pi)\leq\varepsilon$, where $T$ denotes the number of local updates.

\section{Conclusion}
\label{sec:ccl}
We introduced a communication efficient variant of Langevin algorithm for federated learning called Langevin-MARINA. We studied the complexity of this algorithm in terms of KL divergence, Total Variation distance and 2-Wasserstein distance. Unlike existing works on sampling for federated learning, we only require the target distribution to satisfy LSI, which allows the target distribution to not be log-concave. 

Langevin-MARINA is inspired by a optimization algorithm for federated learning called MARINA. MARINA achieves the state of the art in communication complexity among nonconvex optimization algorithms. However, Langevin-MARINA requires more communication rounds than MARINA. The fundamental reason for this is that Langevin-MARINA needs to communicate the Gaussian noise. To solve this issue, one approach is to design a Langevin algorithm allowing for the compression of the Gaussian noise. Another approach is to put the same random seed in each device in order to ensure that they all generate the same $Z_{k+1}$ at step $k$. This approach does not require the communication of the Gaussian noise. We leave this question for future work.

\clearpage
\bibliographystyle{plainnat}
\bibliography{math2}

\clearpage

\newpage
\appendix

\newpage
\tableofcontents
\newpage
\section{New proofs of existing optimization results for MARINA (Algorithm~\ref{alg:capmarina})}

The goal of this section is to we show that the approach that we used to study Langevin-MARINA can be used to study MARINA. 

In particular, we provide a new analysis of MARINA (Algorithm~\ref{alg:capmarina}) establishing the convergence of the whole continuous trajectories $(x_t)_{t\geq 0}$ generated by the algorithm, unlike~\cite{Gorbunov2021} which focuses on the discrete iterates~$(x_k)_{k=0}^{K}$. Compared to~\cite{Gorbunov2021}, we obtain the same convergence rate, but with a different method. 

The approach we take to study MARINA is similar to the approach we took to study Langevin-MARINA: we establish the convergence of the whole continuous trajectories. The key difference is that MARINA is an optmization algorithm in the Euclidean space $\R^d$, so the underlying space in the convergence proofs of MARINA is $\R^d$. On the contrary, we viewed Langevin-MARINA as an optimization algorithm in the Wasserstein space, so the underlying space in the convergence proofs of Langevin-MARINA is the Wasserstein space. 

In this section, we focus on the minimization, by MARINA (Algorithm~\ref{alg:capmarina}), of the empirical risk:
\begin{equation}\label{eq:aaaattt}
	\min_{x\in\R^d} F(x) = \sum_{i=1}^n F_i(x).
\end{equation}


One can see MARINA as a variant of the gradient descent algorithm which uses a MARINA estimator of the gradient $g_k=\frac{1}{n}\sum_{i=1}^n g_k^i$ instead of the full gradient $\nabla F(x_k)$ in the gradient descent algorithm:
\begin{equation}
\label{eq:MARINA_formal}
    x_{k+1} = x_k - h g_{k}.
\end{equation}

MARINA is presented in Algorithm~\ref{alg:capmarina}.





We now provide the convergence results of MARINA. The proofs are provided later in the Appendix.

\begin{theorem}
	\label{thm:mainthmop}
	Let $\left(g_k\right)_{k=0}^{K-1}$ be MARINA estimators. If the step-size satisfies
		\begin{equation}
		\label{eq:stepsizeop}
		0<h\leq\frac{1}{10L}\sqrt{\frac{p}{1+\alpha}},
	\end{equation}
	then
		\begin{equation}
		\label{eq:case1sol2opt}
		\frac{1}{Kh}\int_0^{Kh}\Exp{\normsq{\nabla F(x_t)}}dt\leq\frac{2\left(\Psi_1-F(x^*)\right)}{Kh}+2C\theta,
	\end{equation}
	where $\Psi_1=F(x_0)+hC\G_{0} ,C=\frac{8L^2h^2+2}{1-(1-p)\left(4L^2h^2\alpha+1\right)}$, $x_t:=x_{\lfloor\frac{t}{h}\rfloor h}+(t-\lfloor\frac{t}{h}\rfloor h)x_{\lceil\frac{t}{h}\rceil h}$.
\end{theorem}
Let $h=\frac{1}{10L}\sqrt{\frac{p}{1+\alpha}}$ ~(hence $C=\mathcal{O}\left(\frac{1}{p}\right)$) and $\hat{x}_t:= x_T$, where $T$ is a uniform random variable over $[0,Kh]$ independent of $(x_t)$. Then, $\Exp{\normsq{\nabla F(\hat{x}_t)}} = \frac{1}{Kh}\int_0^{Kh}\Exp{\normsq{\nabla F(x_t)}}dt$. To achieve $\Exp{\normsq{\nabla F(\hat{x}_t)}}\leq\varepsilon^2$ when $\theta=0$ (for instance for the MARINA estimators \eqref{eq:gradientestimator} and  \eqref{eq:gradientestimator222}), $\Omega \left(\frac{(\Psi_1-F(x^*)}{\varepsilon^2}\sqrt{\frac{1+\alpha}{p}}L\right)$ iterations suffice. If $\theta \neq 0$ (for instance for the MARINA estimator~\eqref{eq:gradientestimator333}), if we choose the batch-size $b=\Omega \left(\frac{\sum_{i=1}^n\sigma_i^2}{n^2\varepsilon^2} \right)$, then $2C\theta = \mathcal{O}(\varepsilon^2)$, 
so $\Omega \left(\frac{(\Psi_1-F(x^*))}{\varepsilon^2}\sqrt{\frac{1+\alpha}{p}}L \right)$ iterations suffice.

If $F$ further satisfies the Lojasiewicz condition~\ref{eq:PL}, we can obtain the following stronger result.
\begin{theorem}
	\label{thm:mainthmopawa}
	Let $\left(g_k\right)_{k=0}^{K-1}$ be MARINA estimators. Assume that the gradient domination condition~\ref{eq:PL} holds. If the step-size satisfies
	\begin{equation}
		\label{eq:stepsizeoppl}
		0<h\leq\min \left\{\frac{1}{14L}\sqrt{\frac{p}{1+\alpha}},\frac{p}{6\mu} \right\},
	\end{equation}
	we will have 
	\begin{equation}
		\label{eq:case1sol2optpl}
		\Exp{F(x_{K})}-F(x^*)\leq e^{-\mu Kh}\left(\Psi_2-F(x^*)\right)+\frac{1-e^{-K\mu h}}{\mu}C\theta,
	\end{equation}
	where $\Psi_2=F(x_{0})+\frac{1-e^{-\mu h}}{\mu}C\G_{0} ,C=\frac{8L^2h^2{\beta}+2{\beta}}{1-(1-p)\left(4L^2h^2\alpha+1\right){\beta}}$,$\beta=e^{\mu h}$.
\end{theorem}

Let $h=\min\{\frac{1}{14L}\sqrt{\frac{p}{1+\alpha}},\frac{p}{6\mu}\}$ (hence $C=\mathcal{O}\left(\frac{1}{p}\right)$).  To achieve $\Exp{F(x_{K})}-F(x^*)\leq\varepsilon$ when $\theta=0$ (for instance for the MARINA estimators \eqref{eq:gradientestimator} and  \eqref{eq:gradientestimator222}),
$\Omega(\max\{\sqrt{\frac{1+\alpha}{p}}\frac{L}{\mu},\frac{1}{p}\}\log\left(\frac{\Psi_2-F(x^*)}{\varepsilon})\right)$ iterations suffice. If $\theta \neq 0$ (for instance for the MARINA estimator~\eqref{eq:gradientestimator333}), if we choose the batch-size $b=\Omega\left(\frac{\sum_{i=1}^n\sigma_i^2}{\mu n^2\varepsilon}\right)$, then term $\frac{1-e^{-K\mu h}}{\mu}C\theta$ is of order $\mathcal{O}(\varepsilon)$, so $\Omega\left(\max\{\sqrt{\frac{1+\alpha}{p}}\frac{L}{\mu},\frac{1}{p}\}\log(\frac{\Psi_2-F(x^*)}{\varepsilon})\right)$ iterations suffice.

Since we recover the convergence rates of~\cite{Gorbunov2021}, we refer to the latter paper for a discussion of these results.

\newpage
\section{Proofs}

\subsection{Proof of \Cref{prop:vanilla}}
For gradient estimator \ref{eq:gradientestimator}, we have
\begin{equation}
	\begin{aligned}
		&\Exp{\normsq{g_{k+1}-\nabla F(x_{k+1})}\mid x_{k+1},x_k}\\
		&=(1-p)\Exp{\normsq{g_{k}+\frac{1}{n}\sum_{i=1}^n\mathcal{Q}(\nabla F_{i}(x_{k+1})-\nabla F_i(x_k))-\nabla F(x_{k+1})}\mid x_{k+1},x_k}\\
		&=(1-p)\Exp{\normsq{\frac{1}{n}\sum_{i=1}^n\mathcal{Q}(\nabla F_{i}(x_{k+1})-\nabla F_i(x_k))-\nabla F(x_{k+1})+\nabla F(x_k)}\mid x_{k+1},x_k}\\
		&\quad +(1-p)\Exp{\normsq{g_k-\nabla F(x_k)}\mid x_k}.
	\end{aligned}
\end{equation}
Since $\mathcal{Q}\left(\nabla F_{1}\left(x_{k+1}\right)-\nabla F_{1}\left(x_{k}\right)\right), \ldots, \mathcal{Q}\left(\nabla F_{n}\left(x_{k+1}\right)-\nabla F_{n}\left(x_{k}\right)\right)$ are independent random vectors, now we have 
\begin{equation}
	\begin{aligned}
		&\Exp{\normsq{g_{k+1}-\nabla F(x_{k+1})}\mid x_{k+1},x_k}\\
		&=(1-p)\Exp{\normsq{\frac{1}{n}\sum_{i=1}^n\left(\mathcal{Q}(\nabla F_{i}(x_{k+1})-\nabla F_i(x_k))-\nabla F_i(x_{k+1})+\nabla F_i(x_k)\right)}\mid x_{k+1},x_k}\\
		&\quad +(1-p)\Exp{\normsq{g_k-\nabla F(x_k)}\mid x_k}\\
		&=\frac{1-p}{n^2}\sum_{i=1}^n
		\Exp{\normsq{\mathcal{Q}(\nabla F_{i}(x_{k+1})-\nabla F_i(x_k))-\nabla F_i(x_{k+1})+\nabla F_i(x_k)}\mid x_{k+1},x_k}\\
		&\quad+	(1-p)\Exp{\normsq{g_k-\nabla F(x_k)}\mid x_k}\\
		&\leq \frac{(1-p)\omega}{n^2}\sum_{i=1}^n
		\Exp{\normsq{\nabla F_i(x_{k+1})-\nabla F_i(x_k)}\mid x_{k+1},x_k}+(1-p)\Exp{\normsq{g_k-\nabla F(x_k)}\mid x_k}.
	\end{aligned}
\end{equation}
Use \Cref{eq:Lismooth} and the tower property, we obtain
\begin{equation}
	\begin{aligned}
		&\Exp{\normsq{g_{k+1}-\nabla F(x_{k+1})}}\\
		&=	\Exp{\Exp{\normsq{g_{k+1}-\nabla F(x_{k+1})}\mid x_{k+1},x_k}}\\
		&\leq \frac{(1-p)\omega}{n^2}\sum_{i=1}^n L_i^2 \Exp{\normsq{x_{k+1}-x_k}}+(1-p)\Exp{\normsq{g_k-\nabla F(x_k)}}\\
		&=(1-p)L^2\frac{\omega\sum_{i=1}^n L_i^2}{n^2L^2} \Exp{\normsq{x_{k+1}-x_k}}+(1-p)\Exp{\normsq{g_k-\nabla F(x_k)}},
	\end{aligned}
\end{equation}
so $\alpha=\frac{\omega\sum_{i=1}^n L_i^2}{n^2L^2},\theta=0$.
\subsection{Proof of \Cref{prop:finitesum}}
For gradient estimator \ref{eq:gradientestimator222}~(finite sum case, that is for each $i\in [n]$, $F_i:=\frac{1}{N}\sum_{j=1}^NF_{ij}$), we have
\begin{equation}
	\begin{aligned}
		&\Exp{\normsq{g_{k+1}-\nabla F(x_{k+1})}\mid x_{k+1},x_k}\\
		&=(1-p)\Exp{\normsq{g_k+\frac{1}{n}\sum_{i=1}^n\mathcal{Q}\left(\frac{1}{b'}\sum_{j\in I_{i,k}'}\left(\nabla F_{ij}(x_{k+1})-\nabla F_{ij}(x_k)\right)\right)-\nabla F(x_{k+1})}\mid x_{k+1},x_k}\\
		&=(1-p)\Exp{\normsq{\frac{1}{n}\sum_{i=1}^n\mathcal{Q}\left(\frac{1}{b'}\sum_{j\in I_{i,k}'}\left(\nabla F_{ij}(x_{k+1})-\nabla F_{ij}(x_k)\right)\right)-\nabla F(x_{k+1})+\nabla F(x_k)}\mid x_{k+1},x_k}\\
		&\quad +(1-p)\Exp{\normsq{g_k-\nabla F(x_k)}\mid x_k}.
	\end{aligned}
\end{equation}
Next, we use the notation: $\widetilde{\Delta}_{i}^{k}=\frac{1}{b^{\prime}} \sum_{j \in I_{i, k}^{\prime}}\left(\nabla F_{i j}\left(x_{k+1}\right)-\nabla F_{i j}\left(x_{k}\right)\right)$ and $\Delta_{i}^{k}=\nabla F_{i}\left(x_{k+1}\right)-\nabla F_{i}\left(x_{k}\right)$. They satisfy $\E\left[\widetilde{\Delta}_{i}^{k} \mid x_{k+1}, x_{k}\right]=\Delta_{i}^{k}$ for all $i \in[n]$. Moreover, $\mathcal{Q}\left(\tilde{\Delta}_{1}^{k}\right), \ldots, \mathcal{Q}\left(\tilde{\Delta}_{n}^{k}\right)$ are independent random vectors, now we have
\begin{equation}
	\begin{aligned}
		&\Exp{\normsq{g_{k+1}-\nabla F(x_{k+1})}\mid x_{k+1},x_k}\\
		&=(1-p)\Exp{\normsq{\frac{1}{n}\sum_{i=1}^n\left(\mathcal{Q}(\widetilde{\Delta}_{i}^{k})-\Delta_i^k\right)}\mid x_{k+1},x_k}+(1-p)\Exp{\normsq{g_k-\nabla F(x_k)}\mid x_k}\\
		&=\frac{1-p}{n^2}\sum_{i=1}^n\Exp{\normsq{\mathcal{Q}(\widetilde{\Delta}_{i}^{k})-\widetilde{\Delta}_{i}^{k}+\widetilde{\Delta}_{i}^{k}-\Delta_{i}^k}\mid x_{k+1},x_k}+(1-p)\Exp{\normsq{g_k-\nabla F(x_k)}\mid x_k}\\
		&=\frac{1-p}{n^2}\sum_{i=1}^n\left(\Exp{\normsq{\mathcal{Q}(\widetilde{\Delta}_{i}^{k})-\widetilde{\Delta}_{i}^{k}}\mid x_{k+1},x_k}+\Exp{\normsq{\widetilde{\Delta}_{i}^{k}-\Delta_{i}^k}\mid x_{k+1},x_k}\right)+(1-p)\Exp{\normsq{g_k-\nabla F(x_k)}\mid x_k}\\
		&=\frac{1-p}{n^2}\sum_{i=1}^n\left(\omega
		\Exp{\normsq{\widetilde{\Delta}_{i}^{k}}\mid x_{k+1},x_k}+\Exp{\normsq{\widetilde{\Delta}_{i}^{k}-\Delta_i^k}\mid x_{k+1},x_k}\right)+(1-p)\Exp{\normsq{g_k-\nabla F(x_k)}\mid x_k}\\
		&=\frac{1-p}{n^2}\sum_{i=1}^n\left(\omega
		\Exp{\normsq{{\Delta}_{i}^{k}}\mid x_{k+1},x_k}+(1+\omega)\Exp{\normsq{\widetilde{\Delta}_{i}^{k}-\Delta_i^k}\mid x_{k+1},x_k}\right)+(1-p)\Exp{\normsq{g_k-\nabla F(x_k)}\mid x_k}.
	\end{aligned}
\end{equation}
Next we need to calculate $\Exp{\normsq{\widetilde{\Delta}_{i}^{k}-\Delta_i^k}\mid x_{k+1},x_k}$. For convenience, we will denote $a_{ij}:=\nabla F_{ij}(x_{k+1})-\nabla F_{ij}(x_k)$ and $a_i:=\nabla F_i(x_{k+1})-\nabla F_i(x_k)$. Define 
\begin{equation}
	\chi_{s}=\begin{cases}
		1& \text{with prob.} \frac{1}{N}\\
		2&\text{with prob.} \frac{1}{N}\\
		\quad &\vdots\\
		N&\text{with prob.}\frac{1}{N}
	\end{cases},
\end{equation}
$\{\chi_{s}\}_{s=1}^{b'}$ independent with each other. Let $I_{i,k}'=\bigcup_{s=1}^{b'}\chi_s$, so
\begin{equation}
	\begin{aligned}
		\widetilde{\Delta}_{i}^{k}&=\frac{1}{b^{\prime}} \sum_{j \in I_{i, k}^{\prime}}\left(\nabla F_{i j}\left(x_{k+1}\right)-\nabla F_{i j}\left(x_{k}\right)\right)\\
		&=\frac{1}{b'}\sum_{s=1}^{b'}\sum_{j=1}^N1_{\chi_s=j}a_{ij},
	\end{aligned}
\end{equation}
then
\begin{equation}
	\begin{aligned}
		&\Exp{\normsq{\widetilde{\Delta}_{i}^{k}}-\Delta_{i}^k\mid x_{k+1},x_k}\\
		&=
		\Exp{\normsq{\frac{1}{b'}\sum_{s=1}^{b'}\sum_{j=1}^N1_{\chi_s=j}(a_{ij}-a_i)}\mid x_{k+1},x_k}\\
		&=\frac{1}{b'^2}\left(\sum_{s=1}^{b'}\Exp{\normsq{\sum_{j=1}^N1_{\chi_s=j}(a_{ij}-a_i)}\mid x_{k+1},x_k}\right.\\
		&\left.\quad +\sum_{s\neq s'}\Exp{\inner{\sum_{j=1}^N1_{\chi_s=j}(a_{ij}-a_i)}{\sum_{j=1}^N1_{\chi_s'=j}(a_{ij}-a_i)}\mid x_{k+1},x_k}\right)\\
		&=\frac{1}{b'^2}\left(\sum_{s=1}^{b'}\Exp{\normsq{\sum_{j=1}^N1_{\chi_s=j}(a_{ij}-a_i)}\mid x_{k+1},x_k}\right.\\
		&\left.\quad+\sum_{s\neq s'}{\inner{\Exp{\sum_{j=1}^N1_{\chi_s=j}(a_{ij}-a_i)\mid x_{k+1},x_k}}{\Exp{\sum_{j=1}^N1_{\chi_s'=j}(a_{ij}-a_i)}\mid x_{k+1},x_k}}\right)\\
		&=\frac{1}{b'}\left(\Exp{\normsq{\sum_{j=1}^N1_{\chi_{1}=j}a_{ij}}\mid x_{k+1},x_k}-\normsq{a_i}\right)\\
		&\leq \frac{1}{b'}\Exp{\normsq{\sum_{j=1}^N1_{\chi_{1}=j}a_{ij}}\mid x_{k+1},x_k}\\
		&=\frac{1}{b'}\left(\sum_{j=1}^N\Exp{\normsq{1_{\chi_{1}=j}}}\normsq{a_{ij}}+\sum_{j\neq j'}\Exp{\inner{1_{\chi_{1}=j}}{1_{\chi_{1}=j'}}}\inner{a_{ij}}{a_{ij'}}\right)\\
		&=\frac{1}{b'}\frac{1}{N}\sum_{j=1}^N\normsq{a_{ij}}\\
		&\leq \frac{\mathcal{L}_i^2}{b'}\normsq{x_{k+1}-x_k}.
	\end{aligned}
\end{equation}
Use \Cref{eq:Lismooth} and the tower property, we get 
\begin{equation}\label{eq:lspppp}
	\begin{aligned}
		&\Exp{\normsq{g_{k+1}-\nabla F(x_{k+1})}}\\
		&=	\Exp{\Exp{\normsq{g_{k+1}-\nabla F(x_{k+1})}\mid x_{k+1},x_k}}\\
		&=\frac{1-p}{n^2}\sum_{i=1}^n\left(\omega L_i^2+\frac{(1+\omega)\mathcal{L}_i^2}{b'}\right)\Exp{\normsq{x_{k-1}-x_k}}+(1-p)\Exp{\normsq{g_k-\nabla F(x_k)}}\\
		&=(1-p)L^2\frac{\omega\sum_{i=1}^nL_i^2+(1+\omega)\frac{\sum_{i=1}^n\mathcal{L}^2_i}{b'}}{n^2L^2}\Exp{\normsq{x_{k-1}-x_k}}
		+(1-p)\Exp{\normsq{g_k-\nabla F(x_k)}},
	\end{aligned}
\end{equation}
so $\alpha=\frac{\omega\sum_{i=1}^nL_i^2+(1+\omega)\frac{\sum_{i=1}^n\mathcal{L}^2_i}{b'}}{n^2L^2},\theta=0$.
\subsection{Proof of \Cref{prop:online}}
For gradient estimator \ref{eq:gradientestimator333}~(online case, that is for each $i\in [n]$, $F_i:=\mathbb{E}_{\xi_{i}\sim\mathcal{D}_i}\left[F_{\xi_i}\right]$), we first have 
\begin{equation}
	\begin{aligned}
		&\Exp{\left[\left\|g_{k+1}-\nabla F\left(x_{k+1}\right)\right\|^{2}\mid x_{k+1},x_k\right] }\\
		&=(1-p) \Exp{\left[\left\|g_{k}+\frac{1}{n} \sum_{i=1}^{n} \mathcal{Q}\left(\frac{1}{b^{\prime}} \sum_{\xi \in I_{i, k}^{\prime}}\left(\nabla F_{\xi}\left(x_{k+1}\right)-\nabla F_{\xi}\left(x_{k}\right)\right)\right)-\nabla F\left(x_{k+1}\right)\right\|^{2}\mid x_{k+1},x_k\right]}\\
		&\quad+\frac{p}{n^{2} b^{2}} \Exp{\left[\left\|\sum_{i=1}^{n} \sum_{\xi \in I_{i, k}}\left(\nabla F_{\xi}\left(x_{k+1}\right)-\nabla F\left(x_{k+1}\right)\right)\right\|^{2}\mid x_{k+1}\right]}\\
		&{=}(1-p)\Exp{\left[\left\|\frac{1}{n} \sum_{i=1}^{n} \mathcal{Q}\left(\frac{1}{b^{\prime}} \sum_{\xi \in I_{i, k}^{\prime}}\left(\nabla F_{\xi}\left(x_{k+1}\right)-\nabla F_{\xi}\left(x_{k}\right)\right)\right)-\nabla F\left(x_{k+1}\right)+\nabla F\left(x_{k}\right)\right\|^{2}\mid x_{k+1},x_k\right]}\\
		&\quad+(1-p) \Exp{\left[\left\|g_{k}-\nabla F\left(x_{k}\right)\right\|^{2}\mid x_k\right]}+\frac{p}{n^{2} b^{2}} \sum_{i=1}^{n} \sum_{\xi \in I_{i, k}} \Exp{\left[\left\|\nabla F_{\xi}\left(x_{k+1}\right)-\nabla F\left(x_{k+1}\right)\right\|^{2}\mid x_{k+1}\right]}\\
		&\leq(1-p)\Exp{\left[\left\|\frac{1}{n} \sum_{i=1}^{n} \mathcal{Q}\left(\frac{1}{b^{\prime}} \sum_{\xi \in I_{i, k}^{\prime}}\left(\nabla F_{\xi}\left(x_{k+1}\right)-\nabla F_{\xi}\left(x_{k}\right)\right)\right)-\nabla F\left(x_{k+1}\right)+\nabla F\left(x_{k}\right)\right\|^{2}\mid x_{k+1},x_k\right]}\\
		&\quad+(1-p) \Exp{\left[\left\|g_{k}-\nabla F\left(x_{k}\right)\right\|^{2}\mid x_k\right]}+\frac{p\sum_{i=1}^n\sigma_{i}^2}{n^{2} b},
	\end{aligned}
\end{equation}
here $I'_{i,k}$ consists of $b'$ elements i.i.d. sampled from distribution 
$\mathcal{D}_i$.
In the following, we use the notation: $\widetilde{\Delta}_{i}^{k}=\frac{1}{b^{\prime}} \sum_{\xi \in I_{i, k}^{\prime}}\left(\nabla F_{\xi}\left(x_{k+1}\right)-\nabla F_{\xi}\left(x_{k}\right)\right)$ and $\Delta_{i}^{k}=\nabla F_{i}\left(x_{k+1}\right)-\nabla F_{i}\left(x_{k}\right)$. They satisfy $\E\left[\widetilde{\Delta}_{i}^{k} \mid x_{k+1}, x_{k}\right]=\Delta_{i}^{k}$ for all $i \in[n]$. Moreover, $\mathcal{Q}\left(\tilde{\Delta}_{1}^{k}\right), \ldots, \mathcal{Q}\left(\tilde{\Delta}_{n}^{k}\right)$ are independent random vectors, then we have
\begin{equation}
	\begin{aligned}
		&\Exp{\normsq{g_{k+1}-\nabla F(x_{k+1})}\mid x_{k+1},x_k}\\
		&\leq(1-p)\Exp{\normsq{\frac{1}{n}\sum_{i=1}^n\left(\mathcal{Q}(\widetilde{\Delta}_{i}^{k})-\Delta_i^k\right)}\mid x_{k+1},x_k}+(1-p)\Exp{\normsq{g_k-\nabla F(x_k)}\mid x_k}+\frac{p\sum_{i=1}^n\sigma_{i}^2}{n^{2} b}\\
		&=\frac{1-p}{n^2}\sum_{i=1}^n\Exp{\normsq{\mathcal{Q}(\widetilde{\Delta}_{i}^{k})-\widetilde{\Delta}_{i}^{k}+\widetilde{\Delta}_{i}^{k}-\Delta_{i}^k}\mid x_{k+1},x_k}+(1-p)\Exp{\normsq{g_k-\nabla F(x_k)}\mid x_k}+\frac{p\sum_{i=1}^n\sigma_{i}^2}{n^{2} b}\\
		&=\frac{1-p}{n^2}\sum_{i=1}^n\left(\Exp{\normsq{\mathcal{Q}(\widetilde{\Delta}_{i}^{k})-\widetilde{\Delta}_{i}^{k}}\mid x_{k+1},x_k}+\Exp{\normsq{\widetilde{\Delta}_{i}^{k}-\Delta_{i}^k}\mid x_{k+1},x_k}\right)+(1-p)\Exp{\normsq{g_k-\nabla F(x_k)}\mid x_k}\\
		&\quad +\frac{p\sum_{i=1}^n\sigma_{i}^2}{n^{2} b}\\
		&=\frac{1-p}{n^2}\sum_{i=1}^n\left(\omega
		\Exp{\normsq{\widetilde{\Delta}_{i}^{k}}\mid x_{k+1},x_k}+\Exp{\normsq{\widetilde{\Delta}_{i}^{k}-\Delta_i^k}\mid x_{k+1},x_k}\right)+(1-p)\Exp{\normsq{g_k-\nabla F(x_k)}\mid x_k}\\
		&\quad+\frac{p\sum_{i=1}^n\sigma_{i}^2}{n^{2} b}\\
		&=\frac{1-p}{n^2}\sum_{i=1}^n\left(\omega
		\Exp{\normsq{{\Delta}_{i}^{k}}\mid x_{k+1},x_k}+(1+\omega)\Exp{\normsq{\widetilde{\Delta}_{i}^{k}-\Delta_i^k}\mid x_{k+1},x_k}\right)+(1-p)\Exp{\normsq{g_k-\nabla F(x_k)}\mid x_k}\\
		&\quad+\frac{p\sum_{i=1}^n\sigma_{i}^2}{n^{2} b}.
	\end{aligned}
\end{equation}
Now we need to calculate $\Exp{\normsq{\widetilde{\Delta}_{i}^{k}-\Delta_i^k}\mid x_{k+1},x_k}$. Let $\xi_{i,s}^k\sim\mathcal{D}_i,s=1,2,\ldots,b'$ be $b'$ i.i.d. random variables, $I_{i,k}^{\prime}:=\bigcup_{s=1}^{b'}\xi_{i,s}^k$ and denote $a_{\xi_{i,s}^k}:=\nabla F_{\xi_{i,s}^k}(x_{k+1})-\nabla F_{\xi_{i,s}^k}(x_k),a_i:=\nabla F_i(x_{k+1})-\nabla F_i(x_k)$, then 
\begin{equation}
	\begin{aligned}
		\widetilde{\Delta}_i^k&=\frac{1}{b'}\sum_{\xi \in I_{i, k}^{\prime}}a_{\xi}=\frac{1}{b'}\sum_{k=1}^{b'}a_{\xi_{i,s}^k}.
	\end{aligned}
\end{equation} 
So
\begin{equation}
	\begin{aligned}
		&\Exp{\normsq{\widetilde{\Delta}_{i}^{k}-\Delta_i^k}\mid x_{k+1},x_k}\\
		&=\Exp{\normsq{\frac{1}{b'}\sum_{k=1}^{b'}\left(a_{\xi_{i,s}^k}-a_i\right)}\mid x_{k+1},x_k}\\
		&=\frac{1}{b'^2}\sum_{k=1}^{b'}\Exp{\normsq{a_{\xi_{i,s}^k}-a_i}\mid x_{k+1},x_k}+\frac{1}{b'^2}\sum_{k\neq k'}\Exp{\inner{a_{\xi_{i,s}^k}-a_i}{a_{\chi_{i,k'}}-a_i}\mid x_{k+1},x_k}\\
		&=\frac{1}{b'^2}\sum_{k=1}^{b'}\Exp{\normsq{a_{\xi_{i,s}^k}-a_i}\mid x_{k+1},x_k}\\
		&=\frac{1}{b'^2}\sum_{k=1}^{b'}\Exp{\normsq{a_{\xi_{i,s}^k}-a_i}\mid x_{k+1},x_k}\\
		&=\frac{1}{b'}\left(\Exp{\normsq{a_{\xi_{i,1}^k}}\mid x_{k+1},x_k}-\normsq{a_i}\right)\\
		&\leq\frac{1}{b'}\Exp{\normsq{a_{\xi_{i,1}^k}}\mid x_{k+1},x_k}\\
		&\leq \frac{\mathcal{L}_i^2}{b'}\normsq{x_{k+1}-x_k}.
	\end{aligned}
\end{equation}
Combine with the tower property, we finally get
\begin{equation}
	\begin{aligned}
		&\Exp{\normsq{g_{k+1}-\nabla F(x_{k+1})}}\\
		&=	\Exp{\Exp{\normsq{g_{k+1}-\nabla F(x_{k+1})}\mid x_{k+1},x_k}}\\
		&=\frac{1-p}{n^2}\sum_{i=1}^n\left(\omega L_i^2+\frac{(1+\omega)\mathcal{L}_i^2}{b'}\right)\Exp{\normsq{x_{k+1}-x_k}}+(1-p)\Exp{\normsq{g_k-\nabla F(x_k)}}
		+\frac{p\sum_{i=1}^n\sigma_i^2}{n^2b}\\
		&=(1-p)L^2\frac{\omega\sum_{i=1}^nL_i^2+(1+\omega)\frac{\sum_{i=1}^n\mathcal{L}^2_i}{b'}}{n^2L^2}\Exp{\normsq{x_{k+1}-x_k}}
		+(1-p)\Exp{\normsq{g_k-\nabla F(x_k)}}+\frac{p\sum_{i=1}^n\sigma_i^2}{n^2b},
	\end{aligned}
\end{equation}
so $\alpha=\frac{\omega\sum_{i=1}^nL_i^2+(1+\omega)\frac{\sum_{i=1}^n\mathcal{L}^2_i}{b'}}{n^2L^2},\theta=\frac{p\sum_{i=1}^n\sigma_i^2}{n^2b}$.

\subsection{Proof of Theorem \ref{thm:federatedsampling}}\label{sec:b5}

The following lemma is an integral form of Gr{\"o}nwall inequality from \cite[Chapter II.]{amann2011ordinary}, which plays an important role in the proof of Theorem \ref{thm:federatedsampling} and \ref{thm:mainthmopawa}.

\begin{lemma}[Gr{\"o}nwall Inequality]\label{lem:gronwalllll}
	Assume $\phi, B:[0, T] \rightarrow \mathbb{R}$ are bounded non-negative measurable function and $C:[0, T] \rightarrow \mathbb{R}$ is a non-negative integrable function with the property that
	\begin{equation}
		\label{eq:grrrr1}
		\phi(t) \leq B(t)+\int_{0}^{t} C(\tau) \phi(\tau) d \tau \quad \text { for all } t \in[0, T]
	\end{equation}
	Then
	\begin{equation}
		\label{eq:grrrr2}
		\phi(t) \leq B(t)+\int_{0}^{t} B(s) C(s) \exp \left(\int_{s}^{t} C(\tau) d \tau\right) d s \quad \text { for all } t \in[0, T].
	\end{equation}
\end{lemma}

We also need the following lemma from \cite[Lemma 16.]{chewi2021}.
\begin{lemma}
	\label{lem:Chew}
	Assume that $\nabla F$ is L-Lipschitz. For any probability measure $\mu$, it holds that
	\begin{equation}
		\label{eq:chew}
		\mathbb{E}_{\mu}\left[\|\nabla F\|^{2}\right] \leq \mathbb{E}_{\mu}\left[\left\|\nabla \log( \frac{\mu}{ \pi})\right\|^{2}\right]+2 d L =\FS{\mu}+2dL.
	\end{equation}
\end{lemma}
Remind you the definition of KL divergence and Fisher information:
\begin{equation}
	\KL{\rho_t}:=\int_{\R^d}\log(\frac{\rho_t}{\pi})(x)d\rho_t,\quad\FS{\rho_t}:=\int_{\R^d}\normsq{\nabla \log(\frac{\rho_t}{\pi})}d\rho_t.
\end{equation}
We follow the proof of \cite[Lemma 3]{vempala2019rapid}. Consider the following SDE
\begin{equation}
	dx_t=-f_{\xi}(x_0)dt+\sqrt{2}dB_t,
\end{equation}
where $f_{\cdot}(\cdot): \R^d\times\Xi\to\R^d$, $\Xi$ is of some probability space $\left(\Xi,\rho,\mathcal{F}\right)$, let $\rho_{0t}(x_0,{\xi},x_t)$ denote the joint distribution of $\left(x_0,\xi,x_t\right)$, which we write in terms of the conditionals and marginals as
$$
\rho_{0 t}\left(x_{0},\xi, x_{t}\right)=\rho_{0}\left(x_{0},\xi\right) \rho_{t \mid 0}\left(x_{t} \mid x_{0},\xi\right)=\rho_{t}\left(x_{t}\right) \rho_{0 \mid t}\left(x_{0},\xi \mid x_{t}\right) .
$$
Conditioning on $\left(x_{0},\xi\right)$, the drift vector field $f_{\xi}(x_0)$ is a constant, so the Fokker-Planck formula for the conditional density $\rho_{t \mid 0}\left(x_{t} \mid x_{0}\right)$ is
$$
\frac{\partial \rho_{t \mid 0}\left(x_{t} \mid x_{0},\xi\right)}{\partial t}=\nabla \cdot\left(\rho_{t \mid 0}\left(x_{t} \mid x_{0},\xi\right) f_{\xi}\left(x_{0}\right)\right)+\Delta \rho_{t \mid 0}\left(x_{t} \mid x_{0},\xi \right)
$$
To derive the evolution of $\rho_{t}$, we take expectation over $\left(x_{0},\xi\right) \sim \rho_{0}$, we obtain
\begin{equation}
	\begin{aligned}
		\frac{\partial \rho_{t}(x)}{\partial t} &=\int_{\mathbb{R}^{d}\times\Xi} \frac{\partial \rho_{t \mid 0}\left(x \mid x_{0},\xi\right)}{\partial t} \rho_{0}\left(x_{0},\xi\right) d x_{0}d\xi\\
		&=\int_{\mathbb{R}^{d}\times\Xi}\left(\nabla \cdot\left(\rho_{t \mid 0}\left(x_{t} \mid x_{0},\xi\right) f_{\xi}\left(x_{0}\right)\right)+\Delta \rho_{t \mid 0}\left(x_{t} \mid x_{0},\xi \right)\right) \rho_{0}\left(x_{0},\xi\right) d x_{0}d\xi \\
		&=\int_{\mathbb{R}^{d}\times\Xi}\left(\nabla \cdot\left(\rho_{0t}\left(x, x_{0},\xi\right)  f_{\xi}\left(x_{0}\right)\right)+\Delta \rho_{0t}\left(x, x_{0},\xi\right)\right) d x_{0}d\xi \\
		&=\nabla \cdot\left(\rho_{t}(x) \int_{\mathbb{R}^{d}\times\Xi} \rho_{0 \mid t}\left(x_{0} ,\xi\mid x\right) f_{\xi}\left(x_{0}\right) d x_{0}d\xi\right)+\Delta \rho_{t}(x) \\
		&=\nabla \cdot\left(\rho_{t}(x) \mathbb{E}_{\rho_{0 \mid t}}\left[f_{\xi}\left(x_{0}\right) \mid x_{t}=x\right]\right)+\Delta \rho_{t}(x),
	\end{aligned}
\end{equation}
so we have
\begin{equation}
	\label{eq:tytytyti}
	\begin{aligned}
		\dif{\KL{\rho_t}}&=\int_{\R^d}\left(\nabla \cdot\left(\rho_{t}(x) \mathbb{E}_{\rho_{0 \mid t}}\left[f_{\xi}\left(x_{0}\right) \mid x_{t}=x\right]\right)+\Delta \rho_{t}(x)\right)\log(\frac{\rho_t}{\pi})(x)dx\\
		&=-\int_{\R^d}\inner{\mathbb{E}_{\rho_{0 \mid t}}\left[f_{\xi}\left(x_{0}\right) \mid x_{t}=x\right]+\nabla\log(\rho_t)(x)}{\nabla\log(\frac{\rho_t}{\pi})(x)}\rho_t(x)dx\\
		&=-\int_{\R^d}\inner{\nabla\log(\frac{\rho_t}{\pi})(x)-\nabla\log(\frac{\rho_t}{\pi})(x)+\mathbb{E}_{\rho_{0 \mid t}}\left[f_{\xi}\left(x_{0}\right) \mid x_{t}=x\right]+\nabla\log(\rho_t)(x)}{\nabla\log(\frac{\rho_t}{\pi})(x)}\rho_t(x)dx\\
		&=-\int_{\R^d}\inner{\nabla\log(\frac{\rho_t}{\pi})(x)+\mathbb{E}_{\rho_{0 \mid t}}\left[f_{\xi}\left(x_{0}\right) \mid x_{t}=x\right]-\nabla F(x)}{\nabla\log(\frac{\rho_t}{\pi})(x)}\rho_t(x)dx\\
		&=-\FS{\rho_t}-\int_{\R^d}\inner{\mathbb{E}_{\rho_{0 \mid t}}\left[f_{\xi}\left(x_{0}\right) \mid x_{t}=x\right]-\nabla F(x)}{\nabla\log(\frac{\rho_t}{\pi})(x)}\rho_t(x)dx\\
		&\leq -\FS{\rho_t}+\frac{1}{4}\FS{\rho_t}+\int_{\R^d}\inner{\mathbb{E}_{\rho_{0 \mid t}}\left[f_{\xi}\left(x_{0}\right) \mid x_{t}=x\right]-\nabla F(x)}{\mathbb{E}_{\rho_{0 \mid t}}\left[f_{\xi}\left(x_{0}\right) \mid x_{t}=x\right]-\nabla F(x)}\rho_t(x)dx\\
		&\leq-\frac{3}{4}\FS{\rho_t}+\Exp{\normsq{\mathbb{E}\left[f_{\xi}(x_0)-\nabla F(x_t)\mid x_t\right]}}\\
		&\leq-\frac{3}{4}\FS{\rho_t}+\Exp{\mathbb{E}\left[\normsq{f_{\xi}(x_0)-\nabla F(x_t)}\mid x_t\right]}\\
		&= -\frac{3}{4}\FS{\rho_t}+\Exp{\normsq{f_{\xi}(x_0)-\nabla F(x_t)}}.
	\end{aligned}
\end{equation}
If we replace $f_{\xi}(x_0)$ by $g_k(x_k)$ in \eqref{eq:tytytyti},  we will have
\begin{equation}
	\label{eq:KLM}
	\begin{aligned}
		\dif{\KL{\rho_t}}&\leq -\frac{3}{4}\FS{\rho_t}+\Exp{\normsq{\nabla F(x_t)-g_k}}\\
		&\leq -\frac{3}{4}\FS{\rho_t}+2\Exp{\normsq{\nabla F(x_t)-\nabla F(x_k)}}+2\Exp{\normsq{\nabla F(x_k)-g_k}}\\
		&=-\frac{3}{4}\FS{\rho_t}+2\underbrace{\Exp{\normsq{\nabla F(x_t)-\nabla F(x_k)}}}_{A}+2\G_k.
	\end{aligned}
\end{equation}
We bound term $A$, denote $\cF^k_t$ the filtration generated by $\{B_{s}\}_{s=0}^{kh+t}$, then
\begin{equation}
	\label{eq:temA}
	A=\mathbb{E}_{\rho_t}\left[\mathbb{E}\left[\normsq{\nabla F(x_t)-\nabla F(x_k)}\mid\cF^k_t\right]\right],
\end{equation}
next we estimate the inner expectation,
\begin{equation}
	\label{eq:innerexp}
	\begin{aligned}
		\mathbb{E}\left[\normsq{\nabla F(x_t)-\nabla F(x_k)}\mid\cF^k_t\right]&\leq L^2\mathbb{E}\left[\normsq{ x_t-x_k}\mid\cF^k_t\right]\\
		&=L^2\mathbb{E}\left[\normsq{tg_k+\sqrt{2}\left(B_{kh+t}-B_{kh}\right)}\mid\cF^k_t\right]\\
		&=L^2t^2\normsq{g_k}+2L^2dt\\
		&\leq L^2h^2\normsq{g_k}+2L^2dh\\
		&= L^2\mathbb{E}\left[\normsq{ x_{k+1}-x_k}\mid\cF^k_h\right],
	\end{aligned}
\end{equation}
from \eqref{eq:innerexp} and \eqref{eq:KLM}, we finally have 
\begin{equation}
	\label{eq:finahave}
	\dif{\KL{\rho_t}}\leq -\frac{3}{4}\FS{\rho_t}+2L^2\Exp{\normsq{x_{k+1}-x_k}}+2\G_k.
\end{equation} 
Next, we use Lemma \ref{lem:Chew} to bound $\Exp{\normsq{x_{k+1}-x_k}}$,

\begin{equation}
	\label{eq:bounddif1}
	\begin{aligned}
		\Exp{\normsq{x_{k+1}-x_k}}&=h^2\Exp{\normsq{g_k}}+2dh\\
		&\leq 2h^2\left(\Exp{\normsq{\nabla F(x_k)-g_k}}+\Exp{\normsq{\nabla F(x_k)}}\right)+2dh\\
		& =2h^2 \Exp{\normsq{\nabla F(x_k)}}+2h^2\G_k+2dh\\
		&\leq 4h^2\left(\Exp{\norm{\nabla F(x_t)}}+\Exp{\normsq{\nabla F(x_t)-\nabla F(x_k)}}\right)+2h^2\G_k+2dh\\
		&\leq 4h^2\Exp{\norm{\nabla F(x_t)}}+4L^2h^2\Exp{\normsq{x_{k+1}-x_k}}+2h^2\G_k+2dh,
	\end{aligned}
\end{equation}

so let $h\leq\frac{1}{2\sqrt{2}L}$, we have
\begin{equation}
	\label{eq:kk+1}
	\Exp{\normsq{x_{k+1}-x_k}}\leq 8h^2\Exp{\normsq{\nabla F(x_t)}}+4h^2\G_k+4dh.
\end{equation}
Add $C\G_{k+1}$ to both sides of inequality \eqref{eq:finahave}, $C$ is some constant to be determined later, then use \Cref{def:marina}, the right hand side of \eqref{eq:finahave} will be
\begin{equation}
	\label{eq:Kfinahave}
	\begin{aligned}
		RHS&=-\frac{3}{4}\FS{\rho_t}+2L^2\Exp{\normsq{x_{k+1}-x_k}}+2\G_k+C\G_{k+1}\\
		&\leq-\frac{3}{4}\FS{\rho_t}+2L^2\Exp{\normsq{x_{k+1}-x_k}}+2\G_k+C\left((1-p)\G_k+(1-p)L^2\alpha\Exp{\normsq{x_{k+1}-x_k}}+\theta\right)\\
		&=-\frac{3}{4}\FS{\rho_t}+\left(2L^2+C(1-p)L^2\alpha\right)\Exp{\normsq{x_{k+1}-x_k}}+\left(2+C(1-p)\right)\G_k+C\theta\\
		&\overset{\eqref{eq:kk+1}}{\leq}-\frac{3}{4}\FS{\rho_t}+\left(2L^2+C(1-p)L^2\alpha\right)\left(8h^2\Exp{\normsq{\nabla F(x_t)}}+4h^2\G_k+4dh\right)\\
		&\quad+\left(2+C(1-p)\right)\G_k+C\theta\\
		&\overset{\text{Lemma}~\ref{lem:Chew}}{\leq}-\frac{3}{4}\FS{\rho_t}+\left(2L^2+C(1-p)L^2\alpha\right)\left(8h^2\left(\FS{\mu}+2nL\right)+4h^2\G_k+4dh\right)\\
		&\quad+\left(2+C(1-p)\right)\G_k+C\theta\\
		&=-\left(\frac{3}{4}-8h^2\left(2L^2+C(1-p)L^2\alpha\right)\right)\FS{\rho_t}+\left(8L^2h^2+C(1-p)\left(4L^2h^2\alpha+1\right)+2\right)\G_k\\
		&\quad+\underbrace{\left(2L^2+C(1-p)L^2\alpha\right)\left(8Lh^2d+4dh\right)+C\theta}_{\text{denote as} ~\tau}\\
		&=-\left(\frac{3}{4}-8h^2\left(2L^2+C(1-p)L^2\alpha\right)\right)\FS{\rho_t}+\left(8L^2h^2+C(1-p)\left(4L^2h^2\alpha+1\right)+2\right)\G_k+\tau.
	\end{aligned}
\end{equation}
Choose parameter $\beta$~($\beta=1$ or $\beta=e^{\mu h}$) and let $C=\left(8L^2h^2+C(1-p)\left(4L^2h^2\alpha+1\right)+2\right)\beta$, solve this, we get
\begin{equation}
	\label{eq:newC}
	C=\frac{8L^2h^2\beta+2\beta}{1-(1-p)\left(4L^2h^2\alpha+1\right)\beta}.
\end{equation}

To make sure $C\geq 0$, we should require $h\leq \frac{1}{2L}\sqrt{\frac{p}{(1-p)\alpha}}$ when $\beta=1$, the case $\beta=e^{\mu h}$ is a bit complicated: when $h$ small (for example $h\leq\frac{1}{\mu}$). we have $\beta=e^{\mu h}\leq 1+2\mu h$, insert this into the denominator of $C$ and make sure it positive, that is 
\begin{equation}
	\label{re:eq}
	1-(1-p)\left(4L^2\alpha h^2+1\right)\left(1+2\mu h\right)>0,
\end{equation}
which is equivalent to 
\begin{equation}
	\label{re:ssdsw}
	\underbrace{\frac{1-p}{p}8L^2\mu\alpha h^3}_{I}+\underbrace{\frac{1-p}{p}4L^2\alpha h^2}_{II}+\underbrace{\frac{1-p}{p}2\mu h}_{III}<1,
\end{equation}
one simple solution for \eqref{re:ssdsw} is to let $I<\frac{1}{3},II<\frac{1}{3},III<\frac{1}{3}$, which is 
\begin{equation}\label{re:ree}
	h< \min\{(\frac{p}{24L^2\mu\alpha(1-p)})^{1/3},(\frac{p}{12L^2\alpha (1-p)})^{1/2},\frac{p}{6\mu(1-p)}\}.
\end{equation}
Insert \eqref{eq:newC} into the parameter before $\Exp{\normsq{\nabla F(x_t)}}$ and require
\begin{equation}
	\label{eq:1/22}
	8h^2\left(2L^2+C(1-p)L^2\alpha\right)=8h^2\left(2L^2+\frac{8L^2h^2\beta+2\beta}{1-(1-p)\left(4L^2h^2\alpha+1\right)\beta}(1-p)L^2\alpha\right)\leq\frac{1}{4},
\end{equation}
solve this we get 
\begin{equation}\label{eq:hbounddddd}
	h\leq\frac{1}{2L}\sqrt{\frac{1-(1-p)\beta}{16+(1-p)(17\alpha-16)\beta}}.
\end{equation}
If $\beta=1$, we  need 
\begin{equation}
	\label{eq:hbounuuu}
	{h\leq\frac{1}{10L}\sqrt{\frac{p}{1+\alpha}}}\leq\frac{1}{2L}\sqrt{\frac{p}{17\alpha(1-p)}}\leq\frac{1}{2L}\min\{\sqrt{\frac{p}{16p+17\alpha(1-p)}},\sqrt{\frac{p}{\alpha(1-p)}}\}
\end{equation}
to guarantee $C\geq0$ and \eqref{eq:hbounddddd}.

The  $\beta=e^{\mu h}$ case is complicated: from \eqref{re:ree}, we know $h< \frac{p}{6\mu(1-p)}$, so $\beta=e^{\mu h}\leq 1+2\mu h\leq
1+\frac{p}{3(1-p)}$, insert this upper bound of $\beta$ into \eqref{eq:hbounddddd}, we get a lower bound of the right hand side of \eqref{eq:hbounddddd}, that is 
\begin{equation}
	\label{re:rere}
	\begin{aligned}
		\frac{1}{2L}\sqrt{\frac{2p}{17\alpha(3-2p)+32p}}&=\frac{1}{2L}\sqrt{\frac{1-(1-p)(1+\frac{p}{3(1-p)})}{16+(1-p)(17\alpha-16)(1+\frac{p}{3(1-p)})}}\\
		&\leq\frac{1}{2L}\sqrt{\frac{1-(1-p)\beta}{16+(1-p)(17\alpha-16)\beta}},
	\end{aligned}
\end{equation}
So we need 
{\begin{equation}
		\label{eq:reree}
		h<\min\{\frac{1}{2L}\sqrt{\frac{2p}{17\alpha(3-2p)+32p}},(\frac{p}{24L^2\mu\alpha(1-p)})^{1/3},(\frac{p}{12L^2\alpha (1-p)})^{1/2},\frac{p}{6\mu(1-p)}\}.
\end{equation}}
We can further simplify \eqref{eq:reree}: the first and third term in \eqref{eq:reree} will be greater than $\underbrace{\frac{1}{14L}\sqrt{\frac{p}{1+\alpha}}}_{a}$, the fourth term is greater than $\underbrace{\frac{p}{6\mu}}_{b}$ and $\min\{a,b\}$ is a lower bound of the second term in \eqref{eq:reree}, since $\min\{a,b\}\leq a^{2/3}b^{1/3}=(\frac{p^2}{1176L^2\mu(1+\alpha)})^{1/3}\leq (\frac{p}{24L^2\mu\alpha(1-p)})^{1/3}$. 

So finally when $\beta=e^{\mu h}$, we need
{\begin{equation}
		\label{eq:rererererer}
		h\leq\min\{\frac{1}{14L}\sqrt{\frac{p}{1+\alpha}},\frac{p}{6\mu}\}
\end{equation}}
to guarantee $C\geq0$ and \eqref{eq:hbounddddd}.

Once we have $C\geq 0$ and \eqref{eq:1/22}, then
\begin{equation}
	\label{eq:finnnnn}
	\dif{\KL{\rho_t}}+C\G_{k+1}\leq -\frac{1}{2}\FS{\rho_t}+\beta^{-1}C\G_k+C\tau.
\end{equation}

\textbf{Case I.} $\beta=1$, integrate \eqref{eq:finnnnn}, we have then
\begin{equation}
	\label{eq:nonconvexLA}
	\int_{kh}^{(k+1)h}\FS{\rho_t}dt\leq 2\left(\KL{\rho_{kh}}+C\G_kh-\left(\KL{\rho_{(k+1)h}}+C\G_{k+1}h\right)\right)+C\tau h,
\end{equation}
use \eqref{eq:nonconvexLA} for $k=0,1,2,\cdots,K$, we finally have
{
	\begin{equation}
		\label{eq:nonconvexLAf}
		\begin{aligned}
			\frac{1}{Kh}\int_{0}^{Kh}\FS{\rho_t}dt&\leq \frac{2\left(\KL{\rho_{0}}+C\G_0h-\left(\KL{\rho_{Kh}}+C\G_{K}h\right)\right)}{Kh}+\tau\\
			&\leq \frac{2(\KL{\rho_0}+hC\G_0)}{Kh}+\tau .
		\end{aligned}
\end{equation}}

\textbf{Case II.}
Suppose $\pi$ satisfies LSI with parameter $\mu$, that is
\begin{equation}
	\label{eq:LSI}
	\KL{\nu}\leq \frac{1}{2\mu}\FS{\nu},
\end{equation} 
so with LSI, we have from \eqref{eq:finnnnn} that
\begin{equation}
	\label{eq:LSILA}
	\dif{\KL{\rho_t}}+C\G_{k+1}\leq-\mu\KL{\rho_t}+\beta^{-1}C\G_k+\tau.
\end{equation}
Change \eqref{eq:LSILA} into its equivalent integral form, then it satisfies \eqref{eq:grrrr1} with $\phi(t)=\KL{\rho_t},~B(t)=\left(\beta^{-1}C\G_k-C\G_{k+1}+\tau\right)t+\KL{\rho_{kh}},~C(t)=-\mu$, then by \eqref{eq:grrrr2}, we have
\begin{equation}
	\label{eq;KLKLKL}
	\KL{\rho_t}\leq e^{\mu t}\KL{\rho_{kh}}+\frac{1-e^{-\mu t}}{\mu}\left(\beta^{-1}C\G_k-C\G_{k+1}+\tau\right),
\end{equation}
let $t=h$ and $\beta=e^{\mu h}$, then we have
\begin{equation}
	\label{eq:henhao}
	\begin{aligned}
		\KL{\rho_{(k+1)h}}+\frac{1-e^{-\mu h}}{\mu}C\G_{k+1}&\leq e^{-\mu h}\left(\KL{\rho_{kh}}+e^{\mu h}\frac{1-e^{-\mu h}}{\mu}\beta^{-1}C\G_k\right)+\frac{1-e^{-\mu h}}{\mu}\tau\\
		&=e^{-\mu h}\left(\KL{\rho_{kh}}+\frac{1-e^{-\mu h}}{\mu}C\G_{k}\right)+\frac{1-e^{-\mu h}}{\mu}\tau,
	\end{aligned}
\end{equation}
use \eqref{eq:henhao} for $k=0,1,2,\cdots,K-1$, we have finally
{
	\begin{equation}
		\label{eq:fina666}
		\begin{aligned}
			\KL{\rho_{Kh}}&\leq\KL{\rho_{Kh}}+\frac{1-e^{-\mu h}}{\mu}C\G_{K}\\
			&\leq e^{-\mu Kh}\left(\KL{\rho_0}+\frac{1-e^{-\mu h}}{\mu}C\G_{0}\right)+\frac{1-e^{-K\mu h}}{\mu}\tau,
		\end{aligned}
\end{equation}}
this proves \Cref{thm:federatedsampling}

\subsection{Complexity of Langevin-MARINA}


If  $h\leq\min\{\frac{1}{14L}\sqrt{\frac{p}{1+\alpha}},\frac{p}{6\mu}\}$, we wil have $C=\mathcal{O}(\frac{1}{p})$.  To achieve $\KL{\rho_{K}}\leq e^{-\mu Kh}\Psi_3+\frac{1-e^{-\mu Kh}}{\mu}\tau=\mathcal{O}(\varepsilon)$, we need to bound the residual term $\frac{1-e^{-K\mu h}}{\mu}\tau$ and the contraction term $e^{-\mu K h}\Psi_3$ respectively. When $\theta=0$,  $h=\mathcal{O}\left(\frac{1}{\frac{4L^2(1+\alpha) d}{\mu p\varepsilon}+\sqrt{\frac{8L^3(1+\alpha) d}{\mu
			p\varepsilon}}}\right)=\mathcal{O}\left(\frac{\mu p\varepsilon}{L^2(1+\alpha) d}\right)$ is enough to make  $\frac{1-e^{-K\mu h}}{\mu}\tau=\cO(\varepsilon)$; when $\theta\neq 0$, we need further require $b=\Omega(\frac{\sum_{i=1}^n\sigma_i^2}{\mu n^2\varepsilon})$ to make the left term $\frac{1-e^{-K\mu h}}{\mu}C\theta=\cO(\varepsilon)$. For the contraction term, we need
$K=\Omega\left(\frac{L^2(1+\alpha) d}{\mu^2 p\varepsilon}\log(\frac{\Psi_3}{\varepsilon})\right)$ ~(here we assume $\frac{\mu p\varepsilon}{L^2(1+\alpha) d}\ll \min \left\{\frac{1}{14L}\sqrt{\frac{p}{1+\alpha}},\frac{p}{6\mu} \right\}$) to make $e^{-\mu K h}\Psi_3=\mathcal{O}(\varepsilon)$. So for all gradient estimators, we need $K=\Omega\left(\frac{L^2(1+\alpha) d}{\mu^2 p\varepsilon}\log(\frac{\Psi_3}{\varepsilon})\right)$ and $h=\mathcal{O}\left(\frac{\mu p\varepsilon}{L^2(1+\alpha) d}\right)$ to guarantee $\KL{\rho_{K}}=\mathcal{O}(\varepsilon)$, for algorithm based on \eqref{eq:gradientestimator333}, we need further assume $b=\Omega(\frac{\sum_{i=1}^n\sigma_i^2}{\mu n^2\varepsilon })$.

The analysis of the complexity under the Total Variation distance and the $2$-Wasserstein distance are similar. We want $\normsq{\rho_{K}-\pi}_{TV}\leq \cO(\varepsilon^2)$ and $W_2^2(\rho_{K},\pi)\leq \cO(\varepsilon^2)$,  by \Cref{cor:12} we only need to guarantee $e^{-\mu Kh}\Psi_3+\frac{1-e^{-\mu Kh}}{\mu}\tau=\cO(\varepsilon^2)$ and $e^{-\mu Kh}\Psi_3+\frac{1-e^{-\mu Kh}}{\mu}\tau=\cO (\mu\varepsilon^2)$ respectively. So we only need to replace the $\varepsilon$ in $K,h,b$ in the above by $\varepsilon^2$ and $\mu\varepsilon^2$, then we will have $\norm{\rho_{K}-\pi}_{TV}=\cO(\varepsilon)$ and $W_2(\rho_{K},\pi)=\cO( \varepsilon)$.

\subsection{Proofs of Theorem \ref{thm:mainthmop} and \ref{thm:mainthmopawa}}\label{sec:1and2}

Define two flows, 

\begin{equation}
	\label{eq:exflow11}
	{x}_t={x}_k-tg_k,
\end{equation}
\begin{equation}
	\label{eq:exflow21}
	{x}_t^s=\begin{cases}
		{x}_t & s< t\\
		{x}_t-\int_t^s\nabla F({x}_t^l)dl & s\geq t
	\end{cases},
\end{equation}
where $g_k$ is the gradient estimator of \eqref{eq:gradientestimator}, \eqref{eq:gradientestimator222} or \eqref{eq:gradientestimator333}, $x_0=x_k, x_h=x_{k+1}$, note ${x}_t^s$ is continuous with respect to $s$ and when $s=t$, ${x}_t^s={x}_t$.

Then follow the same procedure as in the last section, we have 

\begin{equation}
	\label{eq:Marina1}
	\begin{aligned}
		\dif{F(x_t)}&=\frac{dF({x}^s_t)}{ds}\mid_{s=t}+\left(\dif{F({x}_t)}-\frac{dF({x}^s_t)}{ds}\mid_{s=t}\right)\\
		&=-\normsq{\nabla F(x_t)}+\langle\nabla F({x}_t),\nabla F({x}_t)-g_k\rangle\\
		&\leq -\normsq{\nabla F({x}_t)}+\frac{1}{4}\normsq{\nabla F({x}_t)}+\normsq{\nabla F({x}_t)-g_k}\\
		&\leq-\frac{3}{4}\normsq{\nabla F(x_t)}+\normsq{\nabla F(x_t)-g_k}\\
		&\leq -\frac{3}{4}\normsq{\nabla F(x_t)}+2\normsq{\nabla F(x_t)-\nabla F(x_k)}+2\normsq{\nabla F(x_k)-g_k}\\
		&\leq -\frac{3}{4}\normsq{\nabla F(x_t)}+2L^2\normsq{x_{k+1}-x_k}+2\normsq{\nabla F(x_k)-g_k},
	\end{aligned}
\end{equation}
since $\normsq{\nabla F(x_t)-\nabla F(x_k)}\leq L^2t^2\normsq{g_k}\leq L^2h^2\normsq{g_k}=L^2\norm{x_{k+1}-x_k}$. We define $\G_k:=\Exp{\normsq{\nabla F(x_k)-g_k}}$, then take expectation on both sides of \eqref{eq:Marina1}, we have
\begin{equation}
	\label{eq:Marinannna}
	\dif{\Exp{F(x_t)}}\leq -\frac{3}{4}\Exp{\normsq{\nabla F(x_t)}}+2L^2\Exp{\normsq{x_{k+1}-x_k}}+2\G_k.
\end{equation}
Next, we bound $\Exp{\normsq{x_{k+1}-x_k}}$,
\begin{equation}
	\label{eq:b1}
	\begin{aligned}
		\Exp{\normsq{x_{k+1}-x_k}}&=h^2\Exp{\normsq{g_k}}\\
		&\leq 2h^2\Exp{\normsq{\nabla F(x_k)}}+2h^2\Exp{\normsq{\nabla F(x_k)-g_k}}\\
		&\leq 4h^2\Exp{\normsq{\nabla F(x_t)}}+4h^2\Exp{\normsq{\nabla F(x_t)-\nabla F(x_k)}}+2h^2\G_k\\
		&\leq 4h^2\Exp{\normsq{\nabla F(x_t)}}+4L^2h^2\Exp{\normsq{x_{k+1}-x_k}}+2h^2\G_k,
	\end{aligned}
\end{equation}
%

which is equivalent to 
\begin{equation}
	\label{eq:bb1}
	\left(1-4L^2h^2\right)\Exp{\normsq{x_{k+1}-x_k}}\leq 4h^2\Exp{\normsq{\nabla F(x_t)}}+2h^2\G_k.
\end{equation}

If we let $h\leq\frac{1}{2\sqrt{2}L}$, then based on \eqref{eq:bb1}, we have
\begin{equation}
	\label{eq:b2}
	\Exp{\normsq{x_{k+1}-x_k}}\leq 8h^2\Exp{\normsq{\nabla F(x_t)}}+4h^2\G_k.		
\end{equation}

Add $C\G_{k+1}$ to both sides of \eqref{eq:Marinannna}, $C$ is some parameter to be determined later.  Then we calculate the right hand side based on ~\Cref{def:marina},
\begin{equation}
	\label{eq:rhs}
	\begin{aligned}
		RHS&=   -\frac{3}{4}\Exp{\normsq{\nabla F(x_t)}}+2L^2\Exp{\normsq{x_{k+1}-x_k}}+2\G_k+C\G_{k+1}\\
		&\leq-\frac{3}{4}\Exp{\normsq{\nabla F(x_t)}}+2L^2\Exp{\normsq{x_{k+1}-x_k}}+2\G_k+C\left((1-p)\G_k+(1-p)L^2\alpha\Exp{\normsq{x_{k+1}-x_k}}+\theta\right)\\
		&\leq -\frac{3}{4}\Exp{\normsq{\nabla F(x_t)}}+\left(2L^2+C(1-p)L^2\alpha\right)\Exp{\normsq{x_{k+1}-x_k}}+\left(2+C(1-p)\right)\G_k+C\theta\\
		&\leq -\frac{3}{4}\Exp{\normsq{\nabla F(x_t)}}+\left(2L^2+C(1-p)L^2\alpha\right)\left(8h^2\Exp{\normsq{\nabla F(x_t)}}+4h^2\G_k\right)+\left(2+C(1-p)\right)\G_k+C\theta\\
		&\leq -\left(\frac{3}{4}-8h^2\left(2L^2+C(1-p)L^2\alpha\right)\right)\Exp{\normsq{\nabla F(x_t)}}+\left(8L^2h^2+C(1-p)\left(4L^2h^2\alpha+1\right)+2\right)\G_k+C\theta.
	\end{aligned}
\end{equation}

Choose $\beta$~(we will set $\beta=1$~or~$\beta=e^{\mu h}$~in the later)~ and let $C=\left(8L^2h^2+C(1-p)\left(4L^2h^2\alpha+1\right)+2\right)\beta$, solve this equation, we get
\begin{equation}
	\label{eq:C}
	C=\frac{8L^2h^2\beta+2\beta}{1-(1-p)\left(4L^2h^2\alpha+1\right)\beta}.
\end{equation}
We need 
\begin{equation}\label{eq:sfskfj}
	C\geq0,\quad \frac{3}{4}-8h^2\left(2L^2+C(1-p)L^2\alpha\right)\geq 
	\frac{1}{2}.
\end{equation}

By similar analysis as in \Cref{sec:b5}, we need require {$h\leq \frac{1}{10L}\sqrt{\frac{p}{1+\alpha}}$} when $\beta=1$ and {$h\leq\min\{\frac{1}{14L}\sqrt{\frac{p}{1+\alpha}},\frac{p}{6\mu}\}$} when $\beta=e^{\mu h}$ to guarantee \eqref{eq:sfskfj}. Once we have \eqref{eq:sfskfj}, then we finally have by \eqref{eq:rhs}
\begin{equation}
	\label{eq:final1}
	\dif{\Exp{F(x_t)}}\leq-\frac{1}{2}\Exp{\normsq{\nabla F(x_t)}}+\beta^{-1}CG_{k}- C\G_{k+1}+C\theta.
\end{equation}

\textbf{Case I.} Let $\beta=1$, then we have 
\begin{equation}
	\label{eq:case1}
	\dif{\Exp{F(x_t)}}+C\G_{k+1}\leq -\frac{1}{2}\Exp{\normsq{\nabla F(x_t)}}+C\G_k+C\theta,
\end{equation}
where $C$ is defined in \eqref{eq:C}, integrating both sides from $0$ to $h$, we have 
\begin{equation}
	\label{case1sol}
	\Exp{F(x_{k+1})}-\Exp{F(x_k)}+hC\G_{k+1}\leq -\frac{1}{2}\int_{0}^{h}\Exp{\normsq{\nabla F(x_t)}}dt+hC\G_k+hC\theta,
\end{equation}
which is equivalent to 
\begin{equation}
	\label{eq:casesol1}
	\int_{0}^{h}\Exp{\normsq{\nabla F(x_t)}}dt\leq 2\left(\left(\Exp{F(x_k)}+hC\G_k\right)-\left(\Exp{F(x_{k+1}}+hC\G_{k+1})\right)\right)+2hC\theta.
\end{equation}
If we first do the same procedure as above for $k=0,1,2,\cdots, K-1$ then take summation from both sides, we will have 
{
	\begin{equation}
		\label{eq:case1sol2}
		\begin{aligned}
			\frac{1}{Kh}\int_0^{Kh}\Exp{\normsq{\nabla F(x_t)}}dt&\leq\frac{2\left(\left(\Exp{F(x_0)}+hC\G_0\right)-\left(\Exp{F(x_{K})}+hC\G_{K}\right)\right)}{Kh}+2C\theta\\
			&\leq \frac{2\left(F(x_0)+hC\G_0-F(x^*)\right)}{Kh}+2C\theta,
		\end{aligned}
	\end{equation} 
}
this proves \Cref{thm:mainthmop}.

\textbf{Case II.}Suppose Lojasiewicz condition holds, that is 
\begin{equation}
	\label{eq:PL1111111}
	\|\nabla F(x)\|^{2} \geq 2 \mu\left(F(x)-\min F\right),\quad\forall x \in \mathbb{R}^{d} .
\end{equation}
Combine \eqref{eq:final1} and \eqref{eq:PL1111111}, we have 
\begin{equation}
	\label{eq:final2121212}
	\dif{\left(\Exp{F(x_t)}-F(x^*)\right)}\leq-\mu \left(\Exp{F(x_t)}-F(x^*)\right)+\beta^{-1}CG_{k}- C\G_{k+1}+C\theta,
\end{equation}
which is equivalent to the integral form 
\begin{equation}
	\label{eq:final2}
	\Exp{F(x_t)}-F(x^*)\leq \left(\beta^{-1}C\G_k- C\G_{k+1}+C\theta\right)t+\Exp{F(x_k)}-F(x^*)+\int_0^t
	-\mu\left(\Exp{F(x_{\tau})}-F(x^*)\right)d\tau,\quad t\in [0,h].
\end{equation}
Now we use Gr{\"o}nwall inequality \ref{lem:gronwalllll}, note \eqref{eq:final2} satisfies \eqref{eq:grrrr1} with $\phi(t)=F(x_t)-F(x^*), B(t)=\left(\beta^{-1}C\G_k- C\G_{k+1}+C\theta\right)t+F(x_k)-F(x^*),C(t)=-\mu$, then by \eqref{eq:grrrr2}, we have
\begin{equation}
	\label{eq:fina4}
	\Exp{F(x_t)}-F(x^*)\leq e^{-\mu t}\left(\Exp{F(x_k)}-F(x^*)\right)+\frac{1-e^{-\mu t}}{\mu}\left(\beta^{-1}C\G_{k}- C\G_{k+1}+C\theta\right),
\end{equation}
let $t=h$ and $\beta=e^{\mu h}$, then we have 
\begin{equation}
	\label{eq:fina5}
	\begin{aligned}
		\Exp{F(x_{k+1})}-F(x^*)+\frac{1-e^{-\mu h}}{\mu}C\G_{k+1}&\leq e^{-\mu h}\left(\Exp{F(x_k)}-F(x^*)+e^{\mu h}\frac{1-e^{-\mu h}}{\mu}\beta^{-1}C\G_k\right)+\frac{1-e^{-\mu h}}{\mu}C\theta\\
		&=e^{-\mu h}\left(\Exp{F(x_{k})}-F(x^*)+\frac{1-e^{-\mu h}}{\mu}C\G_{k}\right)+\frac{1-e^{-\mu h}}{\mu}C\theta,
	\end{aligned}
\end{equation}
use \eqref{eq:fina5} for $k=0,1,2,\cdots,K-1$, we have finally
{
	\begin{equation}
		\label{eq:fina6}
		\begin{aligned}
			\Exp{F(x_{K})}-F(x^*)&\leq\Exp{F(x_{K})}-F(x^*)+\frac{1-e^{-\mu h}}{\mu}C\G_{K}\\
			&\leq e^{-\mu Kh}\left(\Exp{F(x_{0})}-F(x^*)+\frac{1-e^{-\mu h}}{\mu}C\G_{0}\right)+\frac{1-e^{-K\mu h}}{\mu}C\theta\\
			&\leq e^{-\mu K h}\left(F(x_0)+\frac{1-e^{-\mu h}}{\mu}C\G_{0}-F(x^*)\right)+\frac{1-e^{-K\mu h}}{\mu}C\theta,
		\end{aligned}
\end{equation}}
this proves \Cref{thm:mainthmopawa}.

\end{document}